\documentclass[sigconf]{acmart}
\usepackage{multirow}
\usepackage{graphicx}
\usepackage{subcaption}
\usepackage{algorithm}
\usepackage{algpseudocode}
\usepackage{placeins}

\setcopyright{none}
\renewcommand\footnotetextcopyrightpermission[1]{}
\AtBeginDocument{%
  }

\begin{document}
\raggedbottom

\title{GenMatch: An End-to-End Generative Matching Framework for Micro-View Order-Dispatching in Ride-Hailing}

\author{Chuang Liu}
\authornote{First author.}
\affiliation{%
  \institution{Didi Chuxing}
  \city{Hangzhou}
  \country{China}
}
\email{chuangliu@didiglobal.com}

\author{Yuxueqing Zhang}
\affiliation{%
  \institution{Didi Chuxing}
  \city{Beijing}
  \country{China}}
\email{zhangyuxueqing@didiglobal.com}

\author{Tengfei Lyu}
\affiliation{%
  \institution{The Hong Kong University of Science and Technology (Guangzhou)}
  \city{Guangzhou}
  \country{China}}
\email{tlyu077@connect.hkust-gz.edu.cn}

\author{Zirui Yuan}
\affiliation{%
  \institution{The Hong Kong University of Science and Technology (Guangzhou)}
  \city{Guangzhou}
  \country{China}}
\email{zyuan779@connect.hkust-gz.edu.cn}

\author{Weiqi Hu}
\affiliation{%
  \institution{Didi Chuxing}
  \city{Beijing}
  \country{China}
}
\email{huweiqi@didiglobal.com}

\author{Yanghan Cheng}
\affiliation{%
  \institution{Didi Chuxing}
  \city{Beijing}
  \country{China}
}
\email{chengyanghan@didiglobal.com}

\author{Ming Wang}
\affiliation{%
  \institution{Didi Chuxing}
  \city{Beijing}
  \country{China}}
\email{nicholaswangming@didiglobal.com}


\author{Li Ma}
\affiliation{%
  \institution{Didi Chuxing}
  \city{Beijing}
  \country{China}}
\email{malimarey@didiglobal.com}

\author{Zihao Lu}
\authornote{Corresponding author.}
\affiliation{%
  \institution{Didi Chuxing}
  \city{Beijing}
  \country{China}}
\email{luzihao@didiglobal.com}

\renewcommand{\shortauthors}{Liu et al.}

\begin{abstract}
Micro-View Order-Dispatching assigns available drivers to passenger orders within each dispatch batch. It is critical to the service quality and operational efficiency of ride-hailing platforms. Mainstream industrial solutions follow a multi-stage paradigm consisting of model prediction, value calculation, and dispatch matching. Although dispatch quality is determined by the final batch-level assignment, these stages optimize different intermediate objectives. This creates cross-stage objective inconsistency, so improving any single stage does not necessarily improve the overall dispatch result. Generative modeling offers a natural solution by mapping the system context directly to the final output. Motivated by this capability, we formulate Micro-View Order-Dispatching as a generative matching problem and propose an end-to-end \underline{Gen}erative \underline{Match}ing framework (GenMatch), the first generative framework for this task to be deployed in a real-world production environment. However, applying generative modeling to this problem introduces three challenges. First, the model must encode an entire dispatch batch, but each batch forms a dynamic sparse bipartite graph, requiring efficient structured batch-level encoding. Second, replacing the hand-crafted value function requires learning unified business utility from heterogeneous feedback. Third, directly generating an assignment requires tracking the evolving matching state because every selected order-driver pair changes the remaining feasible candidates. GenMatch addresses these challenges through a Context-Aware Bipartite Encoder, a Business-Aware Utility Learner, and a State-Aware Pointer Decoder. Extensive offline evaluations and online A/B tests in five cities across DiDi's international ride-hailing markets demonstrate consistent improvements over competitive baselines, confirming the effectiveness and practicality of GenMatch for industrial order-dispatching.
\end{abstract}




\maketitle
\pagestyle{empty}
\thispagestyle{empty}

\begin{figure}[!htbp]
  \centering
  \includegraphics[width=\linewidth]{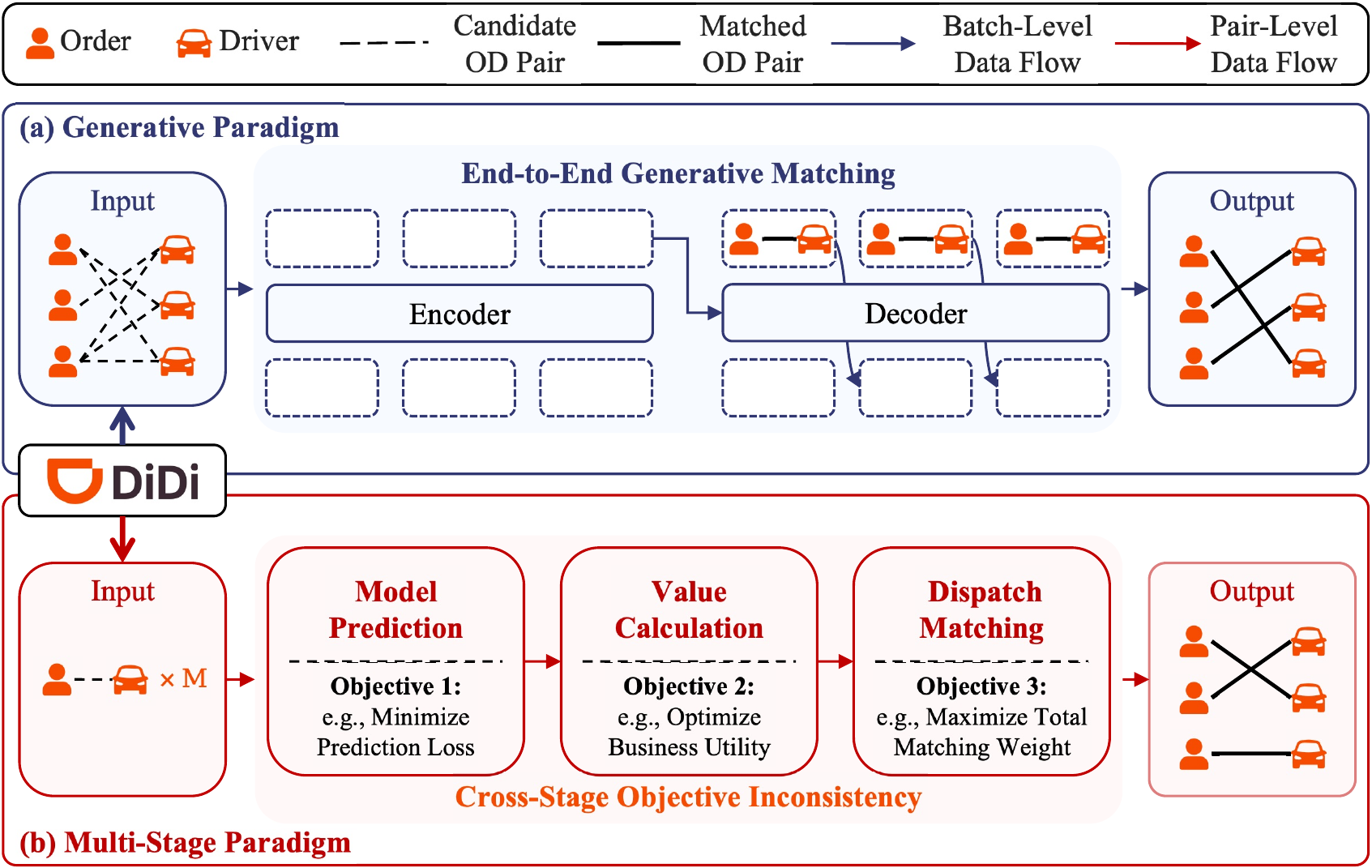}
  \caption{Comparison of \textbf{(a)} the end-to-end generative paradigm and \textbf{(b)} the conventional multi-stage paradigm for Micro-View Order-Dispatching.}
\Description{(a) The proposed end-to-end generative dispatch framework, which models each dispatch batch as a whole and directly generates the dispatch result in an autoregressive manner. (b) The conventional multi-stage dispatch paradigm, which performs pair-level prediction, value calculation, and dispatch matching in separate stages.}
\label{fig:Figue_1}
\end{figure}

\section{Introduction}

Order-dispatching refers to assigning passengers' orders to available drivers in real time. It directly affects passenger and driver experience, which in turn influences platform revenue. Therefore, it is a core process of ride-hailing platforms. Existing studies take two complementary views~\cite{yue2024end}. The Macro-View values and coordinates current decisions under long-term supply--demand evolution~\cite{xu2018large,qin2020ride,tang2021value}. In contrast, the Micro-View considers short-term, real-time assignment within a city. At each fixed dispatch interval, the system collects available orders and drivers; every feasible order-driver combination forms an order-driver (OD) pair, and together they constitute a dispatch batch. Selecting one-to-one assignments from this batch under strict latency constraints is termed \underline{Mic}ro-View \underline{O}rder-\underline{D}ispatching (MICOD).

Mainstream industrial MICOD solutions follow the multi-stage paradigm in Figure~\ref{fig:Figue_1}(b), operating on a single feasible OD pair (pair-level) and the entire dispatch batch (batch-level). They predict pair-level business signals, such as the probabilities of driver answer (DA), passenger cancellation after answer (PCAA), and driver cancellation after answer (DCAA); aggregate them into matching weights with a hand-crafted value function; and apply a batch-level solver such as Kuhn--Munkres matching~\cite{kuhn1955hungarian,munkres1957algorithms}. Although practical, separately optimizing these stages creates \emph{cross-stage objective inconsistency}: better intermediate predictions or weights need not improve the final assignment. Generative models address the same issue by mapping context directly to final outputs in search and query suggestion~\cite{chen2025onesearch,guo2026onesug}, recommendation~\cite{deng2025onerec,zhou2025onerec}, and advertising~\cite{xue2026generative}. We therefore formulate MICOD as end-to-end generation of an assignment from the complete dispatch batch.

This formulation must recover the three capabilities of the replaced pipeline: representing candidates, combining heterogeneous business objectives, and constructing a feasible batch-level matching. This yields three domain-specific challenges.

\noindent\textbf{C1: Structured batch-level encoding of a dynamic sparse bipartite graph.}
Unlike independent pair-level models, a generative model must encode the entire dispatch batch. Its bipartite graph varies in orders, drivers, and feasible OD pairs and is highly sparse, requiring structured batch-level encoding that preserves online computational efficiency.

\noindent\textbf{C2: Learning unified business utility from heterogeneous feedback.}
Replacing manually tuned value rules requires learning business utility directly. Yet generation targets do not fully express outcomes such as DA, PCAA, and DCAA, whose different semantics and directions must be integrated into stable, unified guidance.

\noindent\textbf{C3: Generating batch-level assignments under an evolving matching state.}
Dispatch is evaluated by its joint assignment rather than individual pair scores. During generation, selecting one OD pair invalidates every pair sharing its order or driver and changes the remaining opportunities, requiring an evolving state that supports feasible, batch-coordinated decisions.

To address them, we propose \underline{Gen}erative \underline{Match}ing (GenMatch), which maps a dispatch batch directly to an assignment sequence through the encoder--decoder paradigm in Figure~\ref{fig:Figue_1}(a). Its Context-Aware Bipartite Encoder performs sparse message passing over feasible OD edges; its Business-Aware Utility Learner uses auxiliary supervision to derive business guidance from heterogeneous outcomes; and its State-Aware Pointer Decoder tracks selected and residual candidates while dynamically masking infeasible pairs. Together, they support end-to-end generation under MICOD's structural, business, and matching constraints.

The main contributions are summarized as follows.
\begin{itemize}
    \item To the best of our knowledge, we propose the first end-to-end generative framework for MICOD deployed in a real-world production environment. It directly generates the assignment from an entire dispatch batch and avoids the cross-stage objective inconsistency.
    \item We develop three components for generative matching: a Context-Aware Bipartite Encoder to efficiently encode dynamic sparse bipartite graphs, a Business-Aware Utility Learner to learn unified utility from heterogeneous feedback, and a State-Aware Pointer Decoder to generate feasible assignments under an evolving matching state.
    \item We conduct extensive offline evaluations and online A/B tests in five cities across DiDi's international ride-hailing markets. The results demonstrate consistent improvements over competitive baselines and confirm the effectiveness and practical value of GenMatch in production systems.
\end{itemize}

\section{Related Work} \subsection{Ride-Hailing Order-Dispatching} Effective order-dispatching improves service quality and platform efficiency. Macro-View studies optimize long-term demand--supply dynamics through regional value learning~\cite{xu2018large,tang2019deep,tang2021value}, online or offline reinforcement learning~\cite{sadeghi2022reinforcement,zhang2024nondbrem}, knowledge-enhanced dispatch control~\cite{han2025garlic}, and multi-agent regional cooperation~\cite{yang2024rethinking,jin2019coride,wang2025coopride}.

Micro-View studies make decisions within the current dispatch batch through three paradigms. Policy-based methods such as CoRide and CoopRide produce regional or grid-level actions rather than complete OD-pair assignments~\cite{jin2019coride,wang2025coopride}. D2SN sequentially selects OD pairs or hold actions using a two-layer Markov decision process and an encoder--decoder reinforcement-learning policy~\cite{yue2024end}, but lacks explicit OD-pair interaction modeling and direct supervision from heterogeneous business outcomes. The industrial multi-stage paradigm predicts pair-level signals, calculates matching weights, and applies Greedy~\cite{zheng2018order}, Kuhn--Munkres~\cite{kuhn1955hungarian,munkres1957algorithms}, or Gale--Shapley~\cite{gale1962college} matching. Related systems use deep graph learning for constrained matchmaking and courier pooling~\cite{sun2024mgmatch,liang2024harvesting}, but address generic matchmaking or many-to-one assignment rather than one-to-one MICOD. GenMatch instead unifies batch encoding, business-utility learning, and feasible assignment generation. \subsection{Generative Modeling} Generative modeling produces a target sequence autoregressively from its context instead of scoring only predefined outputs, supporting large or variable output spaces. Industrial systems have applied this formulation to item generation and large-scale sequential recommendation~\cite{rajput2023recommender,zhai2024actions}, food-delivery and e-commerce recommendation~\cite{han2025mtgr,zou2026genrec}, and multi-business, local-life, and advertising scenarios~\cite{li2026mbgr,wei2026oneloc,xue2026generative}. Distributed training systems further scale generative recommendation to industrial workloads~\cite{wang2026mtgenrec}. Recent extensions combine behavioral and semantic information, generate long semantic identifiers in parallel, or apply semantic identifiers to next-point-of-interest recommendation~\cite{wang2024eager,hou2025generating,wang2025generative}. Other studies address the training--inference gap through prefix-aware optimization or combine diffusion with knowledge-graph reasoning~\cite{yu2026apao,zheng2026dikgrec}. Reasoning-augmented language models further support generative next-point-of-interest recommendation~\cite{zhuang2026think2go}. Non-autoregressive generation has also been explored for reranking, while large language models connect quality-aware ranking with candidate generation at web scale~\cite{ren2024non,shah2025towards}.

Generative methods also reduce cross-stage objective inconsistency by replacing retrieval and ranking with direct recommendation generation~\cite{deng2025onerec,zhou2025onerec}, unifying e-commerce search and query suggestion~\cite{chen2025onesearch,chen2026onesearch,guo2026onesug}, or jointly training retrieval and ranking~\cite{li2026unipinrec}. Multi-stage alignment further learns preferences from clicks for generative query suggestion~\cite{yin2026clicks}, while reinforcement learning improves relevance within generative search ranking~\cite{zeng2026optimizing}. Pointer Networks show that structured solutions can be generated by selecting elements from a variable-size input set~\cite{vinyals2015pointer}. MICOD shares this structure, but its candidates form a dynamic sparse bipartite graph whose feasible set changes after every selection. GenMatch extends pointer generation to construct a complete one-to-one matching from such a dispatch batch.

\begin{figure*}[!t]
  \centering
  \includegraphics[width=\linewidth]{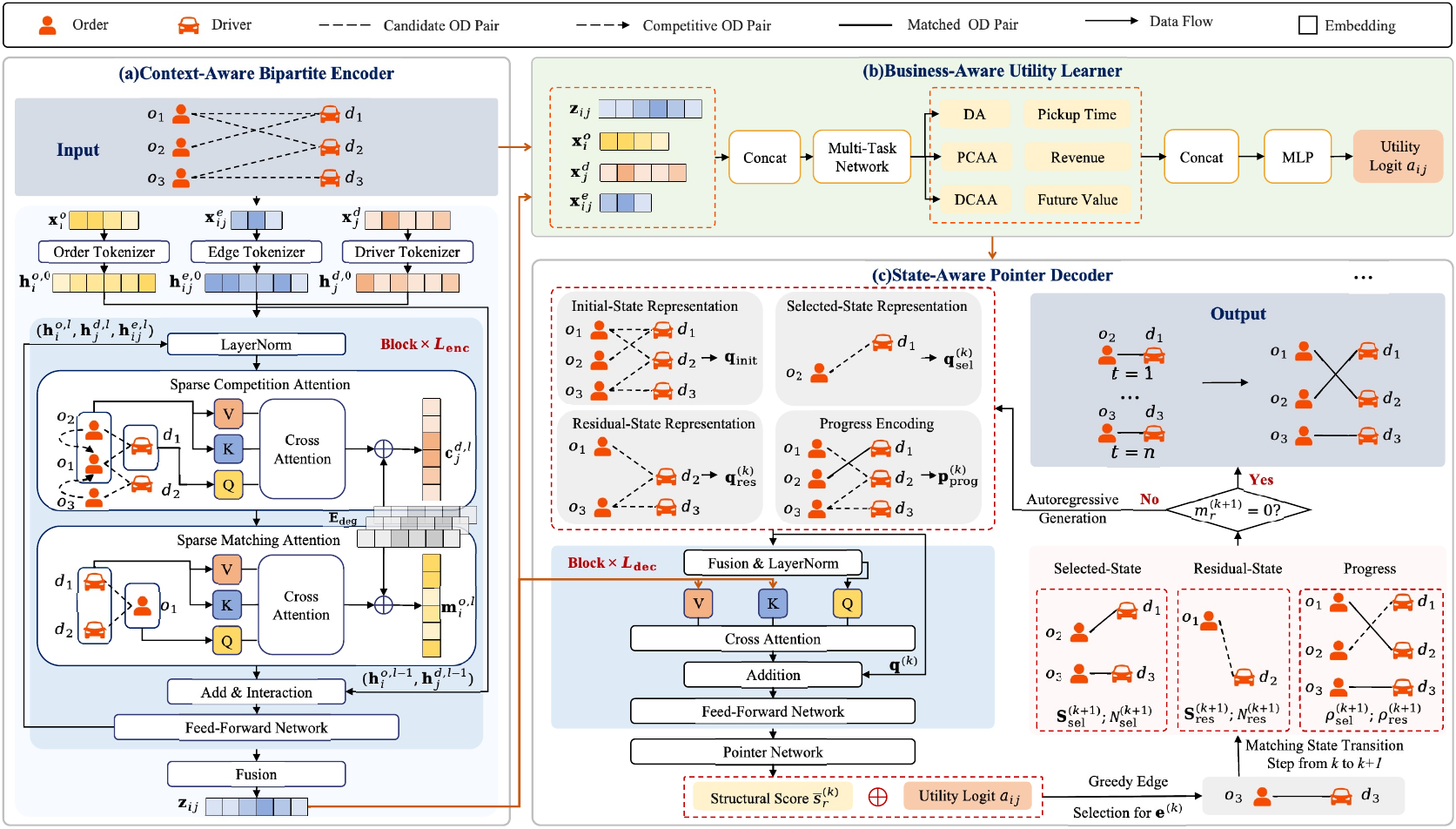}
  \caption{GenMatch architecture: (a) Context-Aware Bipartite Encoder, (b) Business-Aware Utility Learner, and (c) State-Aware Pointer Decoder.}
\Description{The GenMatch architecture consists of a Context-Aware Bipartite Encoder, a Business-Aware Utility Learner, and a State-Aware Pointer Decoder. The decoder updates the selected and residual matching states after generating each OD pair.}
\label{fig:Figue_2}
\end{figure*}

\section{Preliminary}
\label{sec:preliminary}

In this section, we first describe the MICOD process and introduce the corresponding business concepts and notation. We then formulate MICOD as a batch-level matching problem and present its generative formulation.

\noindent\textbf{MICOD Business Process.}
A ride-hailing platform triggers dispatch at fixed intervals. At dispatch step $t$, the platform first collects the currently available passenger orders and drivers. Let $\mathcal{O}_t=\{o_1,\dots,o_{N_t^o}\}$ and $\mathcal{D}_t=\{d_1,\dots,d_{N_t^d}\}$ denote the order and driver sets, respectively. Their sizes vary across dispatch steps. The upstream system then filters infeasible combinations according to pickup distance, service range, and other business rules. The remaining feasible OD pairs form the candidate set $\mathcal{E}_t\subseteq\mathcal{O}_t\times\mathcal{D}_t$. Together, the orders, drivers, and candidate OD pairs constitute a \emph{dispatch batch}, represented as a dynamic sparse bipartite graph
\begin{equation}
\mathcal{G}_t=(\mathcal{O}_t,\mathcal{D}_t,\mathcal{E}_t),
\end{equation}
where each candidate edge $e_{ij}=(o_i,d_j)\in\mathcal{E}_t$ denotes a feasible OD pair. Let $P_t=|\mathcal{E}_t|$ be the number of candidate OD pairs. We index these edges as $\mathcal{E}_t=\{e_1,\ldots,e_{P_t}\}$, where each $e_r$ corresponds to an edge $e_{ij}$. Each order, driver, and candidate edge is associated with a feature vector $\mathbf{x}_i^o\in\mathbb{R}^{d_o}$, $\mathbf{x}_j^d\in\mathbb{R}^{d_d}$, and $\mathbf{x}_{ij}^e\in\mathbb{R}^{d_e}$, respectively. These feature vectors encode spatiotemporal, behavioral, and business signals (e.g., pickup time and platform revenue).

The dispatch system takes the complete dispatch batch $\mathcal{G}_t$ as input and selects a set of OD pairs $\mathcal{M}_t\subseteq\mathcal{E}_t$ for assignment. Each selected pair is broadcast to its driver and enters a stage-wise service process. The driver decides whether to answer. After an answer, the passenger or driver may cancel; the trip is completed only if neither side cancels. This process produces DA, PCAA, DCAA, and trip-completion outcomes. For each broadcast pair $e_{ij}$, $\mathbf{y}_{ij}$ records these events. The multi-task labels follow this service dependency, and downstream events can occur only after driver answer. Unbroadcast candidates have no observed service-process labels and are excluded from the auxiliary loss. Let $\mathcal{C}_t\subseteq\mathcal{M}_t$ denote the set of OD pairs that complete their trips. The completed set $\mathcal{C}_t$ is later used to construct the target sequence for generative training. Candidate retrieval and feasibility filtering are provided by the existing upstream system and are outside the scope of this work.

\noindent\textbf{MICOD Problem Formulation.}
Let $U(\mathcal{M}_t;\mathcal{G}_t)$ denote the business utility of assignment $\mathcal{M}_t$ within dispatch batch $\mathcal{G}_t$. It captures the joint quality of the assignment through outcomes such as DA, PCAA, DCAA, trip completion, pickup time, and platform revenue. Because selected OD pairs compete for shared orders and drivers, this utility is defined over the complete assignment rather than independent pairs. The MICOD objective is to find the feasible matching with the highest batch-level assignment utility:
\begin{equation}
\mathcal{M}_t^*=\arg\max_{\mathcal{M}_t\in\mathcal{F}(\mathcal{G}_t)}U(\mathcal{M}_t;\mathcal{G}_t),
\end{equation}
where $\mathcal{F}(\mathcal{G}_t)$ is the set of feasible matchings over $\mathcal{G}_t$. Each order and each driver can appear in at most one selected OD pair:
\begin{equation}
\begin{aligned}
\sum_{j}\mathbb{I}\!\big((o_i,d_j)\in\mathcal{M}_t\big)&\le 1,
&&\forall o_i\in\mathcal{O}_t,\\
\sum_{i}\mathbb{I}\!\big((o_i,d_j)\in\mathcal{M}_t\big)&\le 1,
&&\forall d_j\in\mathcal{D}_t.
\end{aligned}
\end{equation}
Here, $\mathbb{I}(\cdot)$ is the indicator function.
The matching must also be produced within a strict online latency budget. The conventional multi-stage paradigm approximates this objective by predicting pair-level signals, converting them into matching weights through a hand-crafted value function, and applying a matching solver. In this work, end-to-end means that model prediction, value calculation, and dispatch matching are replaced by one jointly trained model.

\noindent\textbf{Generative MICOD Formulation.}
GenMatch constructs the assignment directly through sequential generation over the candidate OD pairs. An assignment $\mathcal{M}_t$ can be represented by an ordered generation sequence
\begin{equation}
\mathcal{Y}_t=\left(e_t^{(1)},e_t^{(2)},\dots,e_t^{(K_t)}\right),
\end{equation}
where $K_t=|\mathcal{M}_t|$ and the set of generated OD pairs equals $\mathcal{M}_t$. The sequence specifies how the matching is constructed; the final dispatch result is the resulting set rather than a generation order. At generation step $k$, the model selects one candidate $e_t^{(k)}$ from the residual graph $\mathcal{G}_t^{(k)}$. All candidates sharing its order or driver then become infeasible and are removed before the next step. The probability of the complete assignment sequence is factorized as
\begin{equation}
\label{eq:factorization}
\Pr(\mathcal{Y}_t \mid \mathcal{G}_t)=\prod_{k=1}^{K_t}\Pr\!\left(e_t^{(k)} \mid \mathcal{G}_t^{(k)}, e_t^{(<k)}\right),
\end{equation}
where $e_t^{(<k)}$ denotes the selected OD pairs. GenMatch learns to map the complete dispatch batch directly to a feasible assignment while conditioning each decision on the evolving matching state. The next section presents how this formulation is instantiated.

\section{Methods}
\label{sec:methods}

Figure~\ref{fig:Figue_2} presents the Context-Aware Bipartite Encoder, Business-Aware Utility Learner, and State-Aware Pointer Decoder in panels (a), (b), and (c), respectively. They model matching and competition information, learn business-aware candidate representations and utility logits, and generate assignments under the evolving matching state. We follow Section~\ref{sec:preliminary} and omit $t$ when clear; $e_r$ denotes the $r$-th candidate edge, whereas $e_{ij}$ emphasizes its endpoints.

\subsection{Context-Aware Bipartite Encoder}

The suitability of an OD pair depends not only on its own features but also on the other candidates sharing its order or driver. The Context-Aware Bipartite Encoder in Figure~\ref{fig:Figue_2}(a) therefore models batch-level matching and competition relationships over the complete candidate graph. It first summarizes candidates competing at each endpoint, then evaluates potential matches under this competitive context.

\subsubsection{Input Representation}
The encoder takes $\mathcal{G}_t$ and transforms its raw order, driver, and candidate-edge features $\mathbf{x}_i^o$, $\mathbf{x}_j^d$, and $\mathbf{x}_{ij}^e$ with separate tokenizers. The resulting $d$-dimensional embeddings $\mathbf{h}_i^{o,0}$, $\mathbf{h}_j^{d,0}$, and $\mathbf{h}_{ij}^{e,0}$ initialize the sparse bipartite encoder.

\subsubsection{Sparse Matching and Competition Modeling}
The encoder stacks $L_{\mathrm{enc}}$ sparse layers. Layer $l\in\{1,\ldots,L_{\mathrm{enc}}\}$ maps order, driver, and edge states with superscript $l-1$ to those with superscript $l$, first summarizing competitors and then evaluating matches under that context. Let $\mathcal{N}(o_i)$ and $\mathcal{N}(d_j)$ denote the candidate neighbors of order $o_i$ and driver $d_j$.

For each edge $e_{ij}$ incident to driver $d_j$, $\mathbf{r}_{i\rightarrow j}^{l}=\mathbf{h}_i^{o,l-1}+\mathbf{h}_{ij}^{e,l-1}$ carries order and edge information. Let $\mathbf{W}_{q,c}^{d,l}$, $\mathbf{W}_{k,c}^{d,l}$, and $\mathbf{W}_{v,c}^{d,l}$ be the competition-attention projections. We also use a shared degree embedding matrix $\mathbf{E}_{\mathrm{deg}}\in\mathbb{R}^{(D_{\max}+1)\times d}$, where $D_{\max}$ is the maximum retained candidate degree; its indexed row represents neighborhood size. We show one head below, while the implementation concatenates multiple heads. Aggregating $\mathbf{r}_{i\rightarrow j}^{l}$ over $\mathcal{N}(d_j)$ gives competition information $\mathbf{c}_j^{d,l}$:
\begin{equation}
\label{eq:competition-encoding}
\begin{aligned}
\beta_{ij}^{d,l}&=\mathrm{softmax}_{o_i\in\mathcal{N}(d_j)}\!\left(
\frac{(\mathbf{W}_{q,c}^{d,l}\mathbf{h}_j^{d,l-1})^\top
(\mathbf{W}_{k,c}^{d,l}\mathbf{r}_{i\rightarrow j}^{l})}{\sqrt{d}}\right),\\
\mathbf{c}_j^{d,l}&=\sum_{o_i\in\mathcal{N}(d_j)}
\beta_{ij}^{d,l}\mathbf{W}_{v,c}^{d,l}\mathbf{r}_{i\rightarrow j}^{l}
+\mathbf{E}_{\mathrm{deg}}\!\left[|\mathcal{N}(d_j)|\right].
\end{aligned}
\end{equation}
Attention captures neighborhood composition, while the degree embedding preserves its size. Order-side competition information $\mathbf{c}_i^{o,l}$ is computed symmetrically from $\mathbf{r}_{j\rightarrow i}^{l}=\mathbf{h}_j^{d,l-1}+\mathbf{h}_{ij}^{e,l-1}$ over $\mathcal{N}(o_i)$.

We incorporate competition into each candidate representation as $\mathbf{z}_{j\rightarrow i}^{l}=\mathbf{h}_j^{d,l-1}+\mathbf{h}_{ij}^{e,l-1}+\mathbf{c}_j^{d,l}$, with $\mathbf{z}_{i\rightarrow j}^{l}$ defined symmetrically. Using matching-attention projections $\mathbf{W}_{q,m}^{o,l}$, $\mathbf{W}_{k,m}^{o,l}$, and $\mathbf{W}_{v,m}^{o,l}$, their aggregation over drivers produces matching information $\mathbf{m}_i^{o,l}$:
\begin{equation}
\label{eq:matching-encoding}
\begin{aligned}
\alpha_{ij}^{o,l}&=\mathrm{softmax}_{d_j\in\mathcal{N}(o_i)}\!\left(
\frac{(\mathbf{W}_{q,m}^{o,l}\mathbf{h}_i^{o,l-1})^\top
(\mathbf{W}_{k,m}^{o,l}\mathbf{z}_{j\rightarrow i}^{l})}{\sqrt{d}}\right),\\
\mathbf{m}_i^{o,l}&=\sum_{d_j\in\mathcal{N}(o_i)}
\alpha_{ij}^{o,l}\mathbf{W}_{v,m}^{o,l}\mathbf{z}_{j\rightarrow i}^{l}
+\mathbf{E}_{\mathrm{deg}}\!\left[|\mathcal{N}(o_i)|\right].
\end{aligned}
\end{equation}
Driver-side matching information $\mathbf{m}_j^{d,l}$ is computed symmetrically. Competition attention relates candidates sharing an endpoint; matching attention relates a node to its counterparts. Both use $\mathbf{E}_{\mathrm{deg}}$ to preserve neighborhood size and operate only on candidate edges.

Let $\mathrm{LN}$ denote layer normalization, and let $\mathrm{FFN}_{o}^{l}$, $\mathrm{FFN}_{d}^{l}$, and $\mathrm{FFN}_{e}^{l}$ denote the order-, driver-, and edge-side feed-forward networks in layer $l$, respectively. Each block includes its residual connection and layer normalization. The order state is therefore updated as
\begin{equation}
\label{eq:node-update}
\mathbf{h}_i^{o,l}
=\mathrm{FFN}_{o}^{l}\!\left(
\mathbf{h}_i^{o,l-1}+\mathbf{m}_i^{o,l}\right).
\end{equation}
The driver state $\mathbf{h}_j^{d,l}$ is updated symmetrically from $\mathbf{h}_j^{d,l-1}$ and $\mathbf{m}_j^{d,l}$.

The encoder also updates each edge state to retain OD-pair-specific information while incorporating matching information from both endpoints. Let $\odot$ denote element-wise multiplication and $\tilde{\mathbf{h}}_{ij}^{e,l}$ denote the interaction representation. For candidate edge $e_{ij}$, we combine its preceding state with the sum, product, and absolute difference of the order-side and driver-side matching information:
\begin{equation}
\label{eq:edge-update}
\begin{aligned}
\tilde{\mathbf{h}}_{ij}^{e,l}
&=\mathbf{h}_{ij}^{e,l-1}+\mathbf{m}_i^{o,l}+\mathbf{m}_j^{d,l}
+\mathbf{m}_i^{o,l}\odot\mathbf{m}_j^{d,l}
+\left|\mathbf{m}_i^{o,l}-\mathbf{m}_j^{d,l}\right|.
\end{aligned}
\end{equation}
The edge-side block produces $\mathbf{h}_{ij}^{e,l}=\mathrm{FFN}_{e}^{l}(\tilde{\mathbf{h}}_{ij}^{e,l})$. After $L_{\mathrm{enc}}$ layers, the final pair representation is
\begin{equation}
\label{eq:pair-representation}
\mathbf{z}_{ij}=\mathrm{LN}\!\left(
\mathbf{h}_i^{o,L}+\mathbf{h}_j^{d,L}+\mathbf{h}_{ij}^{e,L}
\right).
\end{equation}
The pair representation in Eq.~\eqref{eq:pair-representation} serves as candidate memory for the Business-Aware Utility Learner and State-Aware Pointer Decoder.

\subsection{Business-Aware Utility Learner}

Candidate selection requires reliable representations and explicit business-value guidance. The Business-Aware Utility Learner in Figure~\ref{fig:Figue_2}(b) therefore supervises candidate memory $\mathbf{z}_{ij}$ with stage-wise service outcomes and estimates utility from these outcomes and value signals. These two functions are complementary: auxiliary supervision improves candidate memory and injects service behavior, while the utility logit directly guides subsequent selection.

\subsubsection{Stage-Wise Behavioral Outcome Learning}
Because completion alone cannot identify where an unsuccessful service failed, we use the stage-wise outcomes defined in Section~\ref{sec:preliminary} for fine-grained supervision.

For OD pair $(o_i,d_j)$, let $\mathbf{v}_{ij}=[\mathbf{z}_{ij};\mathbf{x}_i^o;\mathbf{x}_j^d;\mathbf{x}_{ij}^e]$, where $[\,;\,]$ denotes concatenation. A multi-task learning (MTL) network predicts DA, PCAA, and DCAA probabilities $\hat{p}_{ij}^{\mathrm{DA}}$, $\hat{p}_{ij}^{\mathrm{PCAA}}$, and $\hat{p}_{ij}^{\mathrm{DCAA}}$ under their service-stage dependencies. This supervision injects service behavior into $\mathbf{z}_{ij}$.

\subsubsection{Business Utility Estimation}
The stage-wise predictions describe whether an OD pair is likely to become an effective service, but not how much business value it may produce. Actual dispatch decisions must also consider pickup time, platform revenue, future value, and other business signals. Let $\mathbf{b}_{ij}$ denote these business-value fields selected from the raw edge features $\mathbf{x}_{ij}^{e}$. We concatenate them with the three stage-wise predictions as $\mathbf{b}_{ij}^{\mathrm{all}}=[\hat{p}_{ij}^{\mathrm{DA}};\hat{p}_{ij}^{\mathrm{PCAA}};\hat{p}_{ij}^{\mathrm{DCAA}};\mathbf{b}_{ij}]$ and use a multi-layer perceptron (MLP) to estimate the utility logit:
\begin{equation}
\label{eq:business-logit}
a_{ij}=\mathrm{MLP}_{\mathrm{biz}}\!\left(\mathbf{b}_{ij}^{\mathrm{all}}\right).
\end{equation}
The resulting $a_{ij}$ is a candidate-specific and step-independent utility logit. It provides explicit business-value guidance to the subsequent matching process.

\subsection{State-Aware Pointer Decoder}

Each selection changes both feasibility and the remaining matching opportunities. The State-Aware Pointer Decoder in Figure~\ref{fig:Figue_2}(c) therefore conditions on the initial candidates, selected matching, residual candidates, and matching progress. The initial state anchors the original decision space, while the selected and residual states describe committed decisions and remaining opportunities.

\subsubsection{Matching-State Representation}
Because the final assignment is a set, we represent its evolving state with three order-invariant sets: initial, selected, and residual candidates. Let $\{\mathbf{z}_r\}_{r=1}^{P_t}$ be the pair representations and $m_r^{(k)}\in\{0,1\}$ the selectability mask at step $k$, where $m_r^{(k)}=1$ means that $e_r$ remains selectable.

Each matching state is maintained by two state statistics: its representation sum and cardinality. The initial-state statistics are $\mathbf{S}_{\mathrm{init}}=\sum_{r=1}^{P_t}\mathbf{z}_r$ and $N_{\mathrm{init}}=P_t$. If $r^{(s)}$ denotes the candidate selected at step $s$, the selected-state statistics are $\mathbf{S}_{\mathrm{sel}}^{(k)}=\sum_{s<k}\mathbf{z}_{r^{(s)}}$ and $N_{\mathrm{sel}}^{(k)}=k-1$. The residual-state statistics are $\mathbf{S}_{\mathrm{res}}^{(k)}=\sum_{r=1}^{P_t}m_r^{(k)}\mathbf{z}_r$ and $N_{\mathrm{res}}^{(k)}=\sum_{r=1}^{P_t}m_r^{(k)}$.

Dividing each nonempty sum by its cardinality gives $\mathbf{q}_{\mathrm{init}}=\mathbf{S}_{\mathrm{init}}/N_{\mathrm{init}}$, $\mathbf{q}_{\mathrm{sel}}^{(k)}=\mathbf{S}_{\mathrm{sel}}^{(k)}/N_{\mathrm{sel}}^{(k)}$, and $\mathbf{q}_{\mathrm{res}}^{(k)}=\mathbf{S}_{\mathrm{res}}^{(k)}/N_{\mathrm{res}}^{(k)}$, representing the original batch, selected matching, and remaining opportunities. Before any selection, the selected set is empty, so we set $\mathbf{q}_{\mathrm{sel}}^{(k)}$ to a learnable vector $\mathbf{e}_{\mathrm{empty}}$.

Set averaging loses cardinality. We restore it with selected and residual ratios $\rho_{\mathrm{sel}}^{(k)}=N_{\mathrm{sel}}^{(k)}/\min(N_t^o,N_t^d)$ and $\rho_{\mathrm{res}}^{(k)}=N_{\mathrm{res}}^{(k)}/P_t$, normalized by the maximum matching size and initial candidate count. Because batch size is large and dynamic, we scale two learnable $d$-dimensional vectors $\mathbf{e}_{\mathrm{sel}}$ and $\mathbf{e}_{\mathrm{res}}$ rather than use a progress-indexed matrix. The resulting progress encoding and query are
\begin{equation}
\label{eq:state-query}
\begin{aligned}
\mathbf{p}_{\mathrm{prog}}^{(k)} &=
\rho_{\mathrm{sel}}^{(k)}\mathbf{e}_{\mathrm{sel}}+
\rho_{\mathrm{res}}^{(k)}\mathbf{e}_{\mathrm{res}},\\
\mathbf{q}^{(k)} &= \mathrm{LN}\!\left(
\mathbf{W}_{\mathrm{init}}\mathbf{q}_{\mathrm{init}}+
\mathbf{W}_{\mathrm{sel}}\mathbf{q}_{\mathrm{sel}}^{(k)}+
\mathbf{W}_{\mathrm{res}}\mathbf{q}_{\mathrm{res}}^{(k)}+
\mathbf{p}_{\mathrm{prog}}^{(k)}\right).
\end{aligned}
\end{equation}
Here, $\mathbf{W}_{\mathrm{init}}$, $\mathbf{W}_{\mathrm{sel}}$, and $\mathbf{W}_{\mathrm{res}}$ are learnable projections; $\mathbf{p}_{\mathrm{prog}}^{(k)}$ restores the scale information lost by averaging.

\subsubsection{Generative Pointer Distribution}
The query $\mathbf{q}^{(k)}$ is processed by $L_{\mathrm{dec}}$ cross-attention layers over the pair representations, producing the decoder state $\mathbf{h}^{(k)}$. Let $P_{\mathrm{ptr}}$ denote the number of pointer heads, and let $\mathbf{W}_q^{(p)}$ and $\mathbf{W}_k^{(p)}$ be the learnable query and key projections of head $p$. Each head produces a structural matching score $s_r^{(k,p)}$; the final structural score $\bar{s}_r^{(k)}$ is their mean:
\begin{equation}
\label{eq:pointer-score}
\begin{aligned}
s_r^{(k,p)} &=
\frac{(\mathbf{W}_q^{(p)}\mathbf{h}^{(k)})^\top
(\mathbf{W}_k^{(p)}\mathbf{z}_r)}{\sqrt{d}},\\
\bar{s}_r^{(k)} &= \frac{1}{P_{\mathrm{ptr}}}
\sum_{p=1}^{P_{\mathrm{ptr}}}s_r^{(k,p)}.
\end{aligned}
\end{equation}

Let $\mathcal{R}^{(k)}=\{r\mid m_r^{(k)}=1\}$ denote the indices of feasible candidates. For candidate $e_r$, let $a_r$ denote its utility logit; specifically, $a_r=a_{ij}$ when $e_r=e_{ij}$. We denote its conditional generation probability at step $k$ by $p_r^{(k)}$. The decoder combines the step-dependent structural score with this step-independent utility logit:
\begin{equation}
\label{eq:pointer-distribution}
p_r^{(k)}=
\frac{\exp(\bar{s}_r^{(k)}+a_r)}
{\sum_{r'\in\mathcal{R}^{(k)}}\exp(\bar{s}_{r'}^{(k)}+a_{r'})},
\quad r\in\mathcal{R}^{(k)}.
\end{equation}
Here, $a_r$ captures business utility, while $\bar{s}_r^{(k)}$ measures compatibility with the current matching state.

At inference step $k$, the decoder performs greedy edge selection by choosing the candidate with the highest generation probability:
\begin{equation}
\label{eq:greedy-selection}
r^{(k)}=\arg\max_{r\in\mathcal{R}^{(k)}}p_r^{(k)},
\qquad e^{(k)}=e_{r^{(k)}}.
\end{equation}
The selected edge $e^{(k)}$ is appended to the generated assignment, after which the matching state is updated for the next step.

\subsubsection{Matching-State Transition}
After selecting $e^{(k)}=e_{r^{(k)}}$, let the blocked set $\mathcal{B}^{(k)}$ contain it and all selectable pairs sharing its order or driver. The state statistics update incrementally as
\begin{equation}
\label{eq:state-transition}
\begin{aligned}
\mathbf{S}_{\mathrm{sel}}^{(k+1)} &=
\mathbf{S}_{\mathrm{sel}}^{(k)}+\mathbf{z}_{r^{(k)}},\\
N_{\mathrm{sel}}^{(k+1)} &=N_{\mathrm{sel}}^{(k)}+1,\\
\mathbf{S}_{\mathrm{res}}^{(k+1)} &=
\mathbf{S}_{\mathrm{res}}^{(k)}-
\sum_{r\in\mathcal{B}^{(k)}}\mathbf{z}_r,\\
N_{\mathrm{res}}^{(k+1)} &=N_{\mathrm{res}}^{(k)}-|\mathcal{B}^{(k)}|.
\end{aligned}
\end{equation}
We also set $m_r^{(k+1)}=0$ for $r\in\mathcal{B}^{(k)}$ and retain all other mask values, enforcing one-to-one matching.

\subsection{Model Training}

\subsubsection{Target Sequence Construction}
Rather than distill production decisions, we construct supervision from the completed OD-pair set $\mathcal{C}_t$, whose pairs were answered, not canceled, and completed.

Let $K_t^*=|\mathcal{C}_t|$. Because assignment order has no business meaning, each training step uniformly permutes $\mathcal{C}_t$. We denote the resulting target sequence by \mbox{$\mathcal{Y}_t^*=(e_{t,*}^{(1)},\ldots,e_{t,*}^{(K_t^*)})$}. This randomization prevents the decoder from fitting an arbitrary order.

Only completed pairs form targets, but all selectable candidates remain in the denominator of Eq.~\eqref{eq:pointer-distribution} as contrastive alternatives. Thus, GenMatch learns completion patterns from observed targets. At inference, generation continues while $\sum_{r=1}^{P_t}m_r^{(k)}>0$, allowing GenMatch to select unbroadcast candidates exhibiting similar patterns.

\subsubsection{Training Objective}
For assignment generation, we apply teacher forcing and average the negative log-likelihood over all valid target positions in a mini-batch. Let $\mathcal{I}$ denote the dispatch batches in the mini-batch and $Z_{\mathrm{gen}}=\sum_{t\in\mathcal{I}}K_t^*$ the number of valid target positions. The generation loss is
\begin{equation}
\label{eq:generation-loss}
\mathcal{L}_{\mathrm{gen}}=-\frac{1}{Z_{\mathrm{gen}}}
\sum_{t\in\mathcal{I}}\sum_{k=1}^{K_t^*}
\log \Pr\!\left(e_{t,*}^{(k)}\mid
\mathcal{G}_t^{(k)},e_{t,*}^{(<k)}\right).
\end{equation}
For service-process learning, let $\mathcal{T}=\{\mathrm{DA},\mathrm{PCAA},\mathrm{DCAA}\}$ be the auxiliary-task set, and let $\mathrm{BCE}(\cdot,\cdot)$ denote binary cross-entropy. For task $m\in\mathcal{T}$, $y_{tr}^{m}$ is the event label of candidate $e_r$ in batch $t$, and $\hat{p}_{tr}^{m}$ is its predicted event probability. We define $w_{tr}=1$ if $e_r$ was broadcast and its service-process feedback is observable, and $w_{tr}=0$ otherwise. Let $Z_{\mathrm{obs}}=\sum_{t\in\mathcal{I}}\sum_{r=1}^{P_t}w_{tr}$ be the number of candidates with observed feedback. The auxiliary loss is
\begin{equation}
\label{eq:mtl-loss}
\mathcal{L}_{\mathrm{mtl}}=\frac{1}{Z_{\mathrm{obs}}}
\sum_{t\in\mathcal{I}}\sum_{r=1}^{P_t}
w_{tr}
\sum_{m\in\mathcal{T}}\lambda_m
\mathrm{BCE}\!\left(\hat{p}_{tr}^{m},y_{tr}^{m}\right),
\end{equation}
where $\lambda_m$ controls the relative weight of task $m$. Let $\lambda_{\mathrm{mtl}}$ balance service-process learning against assignment generation. The final objective is
\begin{equation}
\label{eq:total-loss}
\mathcal{L} = \mathcal{L}_{\mathrm{gen}}+
\lambda_{\mathrm{mtl}}\mathcal{L}_{\mathrm{mtl}}.
\end{equation}
Both losses are normalized by valid supervision units. We set $\lambda_{\mathrm{mtl}}=10$ based on the sensitivity analysis in Appendix~\ref{sec:appendix_loss_weight}. Appendix~\ref{sec:appendix_procedures} provides the complete training and inference procedures.

\begin{table*}[!t]
  \centering
  \caption{Offline performance relative to PDP$_{\mathrm{KM}}$. Values are the mean\,$\pm$\,standard deviation of percentage changes over five runs; bold and underline denote the best and second-best results.}
  \label{tab:comparison_algorithm}
  \setlength{\tabcolsep}{0.5pt}
  \renewcommand{\arraystretch}{1.1}
  \footnotesize
  \begin{tabular}{@{}l|cccc|cccc|cccc@{}}
    \hline
    \multirow{2}{*}{\textbf{Variant}}
    & \multicolumn{4}{c|}{\textbf{City~I}}
    & \multicolumn{4}{c|}{\textbf{City~II}}
    & \multicolumn{4}{c}{\textbf{City~III}} \\
    & \textbf{AR\,(\%)}\,$\uparrow$ & \textbf{CR\,(\%)}\,$\uparrow$ & \textbf{APT\,(\%)}\,$\downarrow$ & \textbf{GMV\,(\%)}\,$\uparrow$
    & \textbf{AR\,(\%)}\,$\uparrow$ & \textbf{CR\,(\%)}\,$\uparrow$ & \textbf{APT\,(\%)}\,$\downarrow$ & \textbf{GMV\,(\%)}\,$\uparrow$
    & \textbf{AR\,(\%)}\,$\uparrow$ & \textbf{CR\,(\%)}\,$\uparrow$ & \textbf{APT\,(\%)}\,$\downarrow$ & \textbf{GMV\,(\%)}\,$\uparrow$ \\
    \hline
    PDP$_{\mathrm{KM}}$
    & 0.00\,$\pm$\,0.00 & 0.00\,$\pm$\,0.00 & 0.00\,$\pm$\,0.00 & \underline{0.00\,$\pm$\,0.00}
    & 0.00\,$\pm$\,0.00 & \underline{0.00\,$\pm$\,0.00} & 0.00\,$\pm$\,0.00 & 0.00\,$\pm$\,0.00
    & 0.00\,$\pm$\,0.00 & 0.00\,$\pm$\,0.00 & 0.00\,$\pm$\,0.00 & 0.00\,$\pm$\,0.00 \\
    PDP$_{\mathrm{Greedy}}$
    & $-$1.86\,$\pm$\,0.05 & $-$1.69\,$\pm$\,0.09 & $-$0.24\,$\pm$\,0.02 & $-$0.14\,$\pm$\,0.01
    & $-$0.97\,$\pm$\,0.06 & $-$1.26\,$\pm$\,0.04 & \underline{$-$0.25\,$\pm$\,0.01} & $-$1.16\,$\pm$\,0.06
    & $-$1.92\,$\pm$\,0.15 & $-$2.35\,$\pm$\,0.18 & \underline{$-$0.31\,$\pm$\,0.02} & $-$0.19\,$\pm$\,0.02 \\
    PDP$_{\mathrm{GS}}$
    & $-$0.42\,$\pm$\,0.02 & $-$0.30\,$\pm$\,0.03 & $-$0.06\,$\pm$\,0.01 & $-$0.04\,$\pm$\,0.00
    & $-$0.38\,$\pm$\,0.02 & $-$0.54\,$\pm$\,0.02 & $+$0.10\,$\pm$\,0.01 & $-$0.47\,$\pm$\,0.04
    & $-$0.79\,$\pm$\,0.06 & $-$0.76\,$\pm$\,0.07 & $+$0.14\,$\pm$\,0.01 & $-$0.27\,$\pm$\,0.02 \\
    \hline
    D2SN
    & $-$0.09\,$\pm$\,0.03 & $+$0.18\,$\pm$\,0.07 & $-$0.36\,$\pm$\,0.02 & $-$0.05\,$\pm$\,0.00
    & $-$0.03\,$\pm$\,0.09 & $-$0.19\,$\pm$\,0.11 & $-$0.12\,$\pm$\,0.01 & $-$0.47\,$\pm$\,0.06
    & $-$0.37\,$\pm$\,0.12 & $-$0.44\,$\pm$\,0.07 & $-$0.03\,$\pm$\,0.01 & $-$0.24\,$\pm$\,0.04 \\
    RLW
    & $-$0.78\,$\pm$\,0.04 & $-$0.50\,$\pm$\,0.05 & $+$0.42\,$\pm$\,0.02 & $-$0.02\,$\pm$\,0.00
    & $-$0.24\,$\pm$\,0.03 & $-$0.48\,$\pm$\,0.02 & $+$0.34\,$\pm$\,0.03 & $-$0.27\,$\pm$\,0.03
    & $-$0.68\,$\pm$\,0.08 & $-$0.71\,$\pm$\,0.08 & $+$0.22\,$\pm$\,0.03 & $-$0.16\,$\pm$\,0.02 \\
    V1D3
    & $-$1.07\,$\pm$\,0.03 & $-$0.81\,$\pm$\,0.07 & $+$0.29\,$\pm$\,0.01 & $-$0.06\,$\pm$\,0.01
    & $-$0.39\,$\pm$\,0.04 & $-$0.62\,$\pm$\,0.03 & $+$0.26\,$\pm$\,0.02 & $-$0.57\,$\pm$\,0.05
    & $-$0.93\,$\pm$\,0.10 & $-$1.06\,$\pm$\,0.11 & $+$0.18\,$\pm$\,0.02 & $-$0.18\,$\pm$\,0.02 \\
    CoRide
    & $-$1.51\,$\pm$\,0.09 & $-$1.27\,$\pm$\,0.14 & $+$0.98\,$\pm$\,0.03 & $-$0.13\,$\pm$\,0.02
    & $-$0.72\,$\pm$\,0.08 & $-$1.02\,$\pm$\,0.11 & $+$0.67\,$\pm$\,0.03 & $-$0.96\,$\pm$\,0.06
    & $-$2.17\,$\pm$\,0.21 & $-$2.54\,$\pm$\,0.23 & $+$0.47\,$\pm$\,0.04 & $-$0.43\,$\pm$\,0.04 \\
    CoopRide
    & $-$1.13\,$\pm$\,0.07 & $-$0.93\,$\pm$\,0.11 & $+$0.77\,$\pm$\,0.03 & $-$0.10\,$\pm$\,0.01
    & $-$0.55\,$\pm$\,0.06 & $-$0.74\,$\pm$\,0.09 & $+$0.58\,$\pm$\,0.02 & $-$0.76\,$\pm$\,0.05
    & $-$1.77\,$\pm$\,0.14 & $-$1.33\,$\pm$\,0.13 & $+$0.43\,$\pm$\,0.04 & $-$0.33\,$\pm$\,0.03 \\
    \hline
    $\mathbf{GenMatch}_{\mathrm{Value}}$
    & \underline{$+$0.29\,$\pm$\,0.02} & \underline{$+$0.20\,$\pm$\,0.03} & \textbf{$-$0.76\,$\pm$\,0.02} & $-$0.02\,$\pm$\,0.00
    & \underline{$+$0.14\,$\pm$\,0.02} & $-$0.09\,$\pm$\,0.01 & $-$0.19\,$\pm$\,0.01 & \underline{$+$0.03\,$\pm$\,0.01}
    & \underline{$+$0.46\,$\pm$\,0.05} & \underline{$+$0.67\,$\pm$\,0.06} & \textbf{$-$0.58\,$\pm$\,0.05} & \underline{$+$0.24\,$\pm$\,0.02} \\
    \textbf{GenMatch}
    & \textbf{$+$0.51\,$\pm$\,0.03} & \textbf{$+$0.62\,$\pm$\,0.04} & \underline{$-$0.72\,$\pm$\,0.03} & \textbf{$+$0.11\,$\pm$\,0.01}
    & \textbf{$+$0.31\,$\pm$\,0.03} & \textbf{$+$0.23\,$\pm$\,0.02} & \textbf{$-$0.40\,$\pm$\,0.03} & \textbf{$+$0.23\,$\pm$\,0.03}
    & \textbf{$+$0.83\,$\pm$\,0.07} & \textbf{$+$1.17\,$\pm$\,0.12} & $-$0.23\,$\pm$\,0.03 & \textbf{$+$0.55\,$\pm$\,0.03} \\
    \hline
  \end{tabular}
\end{table*}

\section{Experiments}

\subsection{Experimental Settings}
\noindent\textbf{Datasets.}
The experiments cover five cities across DiDi's international ride-hailing markets. Offline evaluation uses City~I--III, while online A/B tests use City~III--V; City~III appears in both settings. The upstream retrieval system is fixed, so all methods receive candidate sets constructed under the same feasibility rules. Detailed city statistics are provided in Appendix~\ref{sec:appendix_experimental_details}.

\noindent\textbf{Offline evaluation} uses simulation environments constructed from historical dispatch logs, with 14 days for training and 7 days for evaluation. Each simulator replays real order requests, driver states, and candidate connections and advances according to the generated decisions.

\noindent\textbf{Online A/B tests} run under live production traffic for 14 days using a 1-hour time-slice interleaved design. The Production Dispatching Pipeline (PDP) is the deployed multi-stage baseline; its Kuhn--Munkres variant, denoted PDP$_{\mathrm{KM}}$, serves as the control. Results are reported as treatment--control (T--C) deltas.

\noindent\textbf{Metrics.}
We use Answer Ratio (AR), Completion Ratio (CR), Average Pickup Time (APT), and Gross Merchandise Volume (GMV). Higher AR, CR, and GMV and lower APT indicate better performance. Detailed metric definitions are provided in Appendix~\ref{sec:appendix_experimental_details}.
\noindent\textbf{Baselines.}
We compare GenMatch with PDP$_{\mathrm{KM}}$ and two solver variants, PDP$_{\mathrm{Greedy}}$ and PDP$_{\mathrm{GS}}$, where GS denotes Gale--Shapley matching~\cite{gale1962college}; end-to-end MICOD method D2SN~\cite{yue2024end}; value-based methods V1D3~\cite{tang2021value} and RLW~\cite{sadeghi2022reinforcement}; and multi-agent methods CoRide~\cite{jin2019coride} and CoopRide~\cite{wang2025coopride}. We also evaluate GenMatch$_{\mathrm{Value}}$ to isolate pair-level prediction. Detailed definitions are provided in Appendix~\ref{sec:appendix_experimental_details}.
\noindent\textbf{Implementation.}
All offline results are averaged over five independent runs. Hardware, optimization, and model configurations are provided in Appendix~\ref{sec:appendix_model_config}.

\subsection{Offline Performance}

Table~\ref{tab:comparison_algorithm} shows that GenMatch consistently outperforms the production and research baselines. Relative to PDP$_{\mathrm{KM}}$, it improves AR, CR, and GMV by 0.31\%--0.83\%, 0.23\%--1.17\%, and 0.11\%--0.55\%, respectively, while reducing APT by 0.23\%--0.72\%. D2SN also generates assignments sequentially but remains inferior on every metric, indicating that generation alone is insufficient without explicit batch interactions, business guidance, and supervised outcome learning. The value-based and multi-agent baselines likewise cannot consistently improve the current batch assignment.

GenMatch$_{\mathrm{Value}}$, which replaces only PDP's prediction stage, improves most metrics and validates the learned service-process predictions. Full GenMatch further improves AR, CR, and GMV by replacing value calculation and matching with state-aware generation, directly supporting our cross-stage inconsistency motivation. Appendix~\ref{sec:appendix_auxiliary_prediction} further evaluates the auxiliary predictions.

\begin{table}[!t]
  \centering
  \caption{Core City~III ablations relative to GenMatch (Full), reported as the mean\,$\pm$\,standard deviation of percentage changes over five runs. Bold denotes the best result in each column.}
  \label{tab:ablation_study}
  \setlength{\tabcolsep}{3.5pt}
  \renewcommand{\arraystretch}{1.08}
  \resizebox{\columnwidth}{!}{%
  \begin{tabular}{@{}llrrrr@{}}
    \hline
    \textbf{Module} & \textbf{Variant} & \textbf{AR}$\uparrow$ & \textbf{CR}$\uparrow$ & \textbf{APT}$\downarrow$ & \textbf{GMV}$\uparrow$ \\
    \hline
    \textbf{GenMatch} & \textbf{Full} & \textbf{0.00\,$\pm$\,0.00} & \textbf{0.00\,$\pm$\,0.00} & \textbf{0.00\,$\pm$\,0.00} & \textbf{0.00\,$\pm$\,0.00} \\
    \hline
    \multirow{3}{*}{Encoder} & A1 & $-$2.55\,$\pm$\,0.12 & $-$3.07\,$\pm$\,0.07 & $+$1.16\,$\pm$\,0.02 & $-$2.28\,$\pm$\,0.15 \\
    & A2 & $-$2.43\,$\pm$\,0.06 & $-$2.69\,$\pm$\,0.03 & $+$1.14\,$\pm$\,0.01 & $-$2.15\,$\pm$\,0.06 \\
    & A3 & $-$0.39\,$\pm$\,0.06 & $-$0.25\,$\pm$\,0.03 & $+$0.18\,$\pm$\,0.02 & $-$0.40\,$\pm$\,0.07 \\
    \hline
    \multirow{2}{*}{Learner} & A4 & $-$2.76\,$\pm$\,0.16 & $-$3.15\,$\pm$\,0.08 & $+$1.17\,$\pm$\,0.03 & $-$2.32\,$\pm$\,0.18 \\
    & A5 & $-$2.02\,$\pm$\,0.08 & $-$2.34\,$\pm$\,0.04 & $+$1.53\,$\pm$\,0.02 & $-$1.88\,$\pm$\,0.09 \\
    \hline
    \multirow{3}{*}{Decoder} & A8 & $-$1.11\,$\pm$\,0.04 & $-$1.07\,$\pm$\,0.02 & $+$0.86\,$\pm$\,0.01 & $-$1.45\,$\pm$\,0.05 \\
    & A9 & $-$1.22\,$\pm$\,0.14 & $-$0.96\,$\pm$\,0.09 & $+$0.48\,$\pm$\,0.04 & $-$0.78\,$\pm$\,0.19 \\
    & A12 & $-$2.46\,$\pm$\,0.15 & $-$3.07\,$\pm$\,0.20 & $+$0.16\,$\pm$\,0.03 & $-$0.91\,$\pm$\,0.05 \\
    \hline
  \end{tabular}}
\end{table}

\subsection{Ablation Study}
\label{sec:ablation}

Table~\ref{tab:ablation_study} reports core ablations on City~III, organized by module to test whether each targeted design contributes to the final assignment. For the encoder, A1 removes batch-level matching and competition information, while A2 restores matching information alone. A2 improves AR, CR, and GMV over A1, and full GenMatch further improves all metrics, validating both forms of context. Removing the shared degree embedding (A3) also degrades every metric, showing that neighborhood size complements attention-based neighbor composition.

For the learner, removing auxiliary supervision (A4) causes the largest AR and CR drops and reduces GMV by 2.32\%, while removing the utility logit (A5) produces the largest APT increase; both representation supervision and business guidance are therefore necessary. For the decoder, removing residual-state or progress information (A8--A9) consistently hurts performance. A12 uses the first-step generation logits as fixed Kuhn--Munkres weights and reduces AR and CR by 2.46\% and 3.07\%, showing that static scores cannot replace state-aware sequential generation. Appendix~\ref{sec:appendix_full_ablation} reports the remaining variants and complete three-city results.

\begin{table*}[t]
  \centering
  \caption{Online A/B test improvements over PDP$_{\mathrm{KM}}$ (T$-$C). Overall averages the three cities; $^{*}$ indicates $p<0.05$.}
  \label{tab:AB_testing}
  \setlength{\tabcolsep}{2pt}
  \renewcommand{\arraystretch}{1.1}
  \resizebox{\textwidth}{!}{
  \footnotesize
  \begin{tabular}{l|cccc|cccc|cccc|cccc}
    \hline
    \multirow{2}{*}{\textbf{Variant}}
    & \multicolumn{4}{c|}{\textbf{City~III}}
    & \multicolumn{4}{c|}{\textbf{City~IV}}
    & \multicolumn{4}{c|}{\textbf{City~V}}
    & \multicolumn{4}{c}{\textbf{Overall}} \\
    & \textbf{AR}\,$\uparrow$ & \textbf{CR}\,$\uparrow$ & \textbf{APT}\,$\downarrow$ & \textbf{GMV}\,$\uparrow$
    & \textbf{AR}\,$\uparrow$ & \textbf{CR}\,$\uparrow$ & \textbf{APT}\,$\downarrow$ & \textbf{GMV}\,$\uparrow$
    & \textbf{AR}\,$\uparrow$ & \textbf{CR}\,$\uparrow$ & \textbf{APT}\,$\downarrow$ & \textbf{GMV}\,$\uparrow$
    & \textbf{AR}\,$\uparrow$ & \textbf{CR}\,$\uparrow$ & \textbf{APT}\,$\downarrow$ & \textbf{GMV}\,$\uparrow$ \\
    \hline
    $\mathbf{GenMatch}_{\mathrm{Value}}$
    & 0.88\% & 2.31\%$^{*}$ & \textbf{$\mathbf{-}$7.08\%$^{*}$} & 2.35\%$^{*}$
    & 0.59\% & 3.06\%$^{*}$ & \textbf{$\mathbf{-}$4.55\%$^{*}$} & 2.43\%$^{*}$
    & 0.75\%$^{*}$ & 1.67\%$^{*}$ & \textbf{$\mathbf{-}$4.05\%$^{*}$} & 1.04\%
    & 0.77\%$^{*}$ & 1.93\%$^{*}$ & \textbf{$\mathbf{-}$4.85\%$^{*}$} & 1.49\%$^{*}$ \\
    \textbf{GenMatch}
    & \textbf{3.18\%$^{*}$} & \textbf{5.37\%$^{*}$} & $-$2.20\%$^{*}$ & \textbf{4.89\%$^{*}$}
    & \textbf{1.72\%$^{*}$} & \textbf{4.51\%$^{*}$} & $-$1.68\%$^{*}$ & \textbf{3.93\%$^{*}$}
    & \textbf{2.01\%$^{*}$} & \textbf{3.26\%$^{*}$} & $-$1.76\%$^{*}$ & \textbf{2.16\%$^{*}$}
    & \textbf{2.26\%$^{*}$} & \textbf{3.86\%$^{*}$} & $\mathbf{-}$1.84\%$^{*}$ & \textbf{2.97\%$^{*}$} \\
    \hline
  \end{tabular}}
\end{table*}

\subsection{Online A/B Testing}
\label{sec:online_ab}

\noindent\textbf{Overall online performance.}
Table~\ref{tab:AB_testing} shows significant improvements in every city. Overall, GenMatch increases AR, CR, and GMV by 2.26\%, 3.86\%, and 2.97\%, respectively, while reducing APT by 1.84\%. GenMatch$_{\mathrm{Value}}$ also improves all four metrics over PDP$_{\mathrm{KM}}$, showing that the service-process predictions learned by GenMatch provide more effective pair-level inputs to the production pipeline. Full GenMatch further improves AR, CR, and GMV over GenMatch$_{\mathrm{Value}}$ by 1.49\%, 1.93\%, and 1.48\% overall. The offline and online results therefore show the same pattern: better pair-level predictions help, while end-to-end generation yields the strongest overall gains. This consistency supports our cross-stage objective inconsistency motivation and shows that the offline optimization transfers to live traffic.

\begin{table}[!htbp]
  \centering
  \caption{Online changes in dispatch effectiveness and experience relative to PDP$_{\mathrm{KM}}$ (T$-$C); $^{*}$ indicates $p<0.05$.}
  \label{tab:experience_metrics}
  \setlength{\tabcolsep}{6pt}
  \renewcommand{\arraystretch}{1.2}
  \footnotesize
  \begin{tabular}{lc}
    \hline
    \textbf{Metric} & \textbf{Delta (T$-$C)} \\
    \hline
    \multicolumn{2}{l}{\textbf{Dispatch Effectiveness Measures}} \\
    Broadcast Count & $-$0.17\%$^{*}$ \\
    Answer Count\,($\uparrow$) & 2.16\%$^{*}$ \\
    Completion Count\,($\uparrow$) & 3.84\%$^{*}$ \\
    \hline
    \multicolumn{2}{l}{\textbf{Passenger Experience Measures}} \\
    Passenger Bad Experience Ratio (PBE)\,($\downarrow$) & $-$15.17\%$^{*}$    \\
    Passenger Cancel Before Answer Ratio (PCBA)\,($\downarrow$) & $-$9.28\%$^{*}$    \\
    Passenger Cancel After Answer Ratio (PCAA)\,($\downarrow$) & $-$7.61\%$^{*}$    \\
    \hline
    \multicolumn{2}{l}{\textbf{Driver Experience Measures}} \\
    Driver Income\,($\uparrow$) & 2.99\%$^{*}$   \\
    Driver Answer Ratio (DA)\,($\uparrow$) &  13.96\%$^{*}$   \\
    Driver Cancel After Answer Ratio (DCAA)\,($\downarrow$) & $-$6.99\%$^{*}$   \\
    \hline
  \end{tabular}
\end{table}

\noindent\textbf{Dispatch effectiveness and user experience.}
Table~\ref{tab:experience_metrics} provides a finer-grained view of the dispatch funnel and service quality. Broadcast Count decreases by 0.17\%, while Answer Count and Completion Count increase by 2.16\% and 3.84\%, respectively. This result is consistent with the greedy autoregressive policy of GenMatch: it suppresses broadcasts that are unlikely to yield an answer or completion and generates more effective assignments. All passenger and driver experience metrics also improve. PBE decreases by 15.17\%, PCBA and PCAA decrease by 9.28\% and 7.61\%, Driver Income and DA increase by 2.99\% and 13.96\%, and DCAA decreases by 6.99\%. Passenger and driver willingness is highly uncertain in international markets. The conventional paradigm predicts these signals separately and then combines them through a hand-crafted value function, so errors can propagate across stages. GenMatch instead learns from heterogeneous behavioral feedback and uses it to guide the final assignment directly, improving both dispatch effectiveness and user experience.

\begin{figure}[t]
  \centering
  \includegraphics[width=\linewidth]{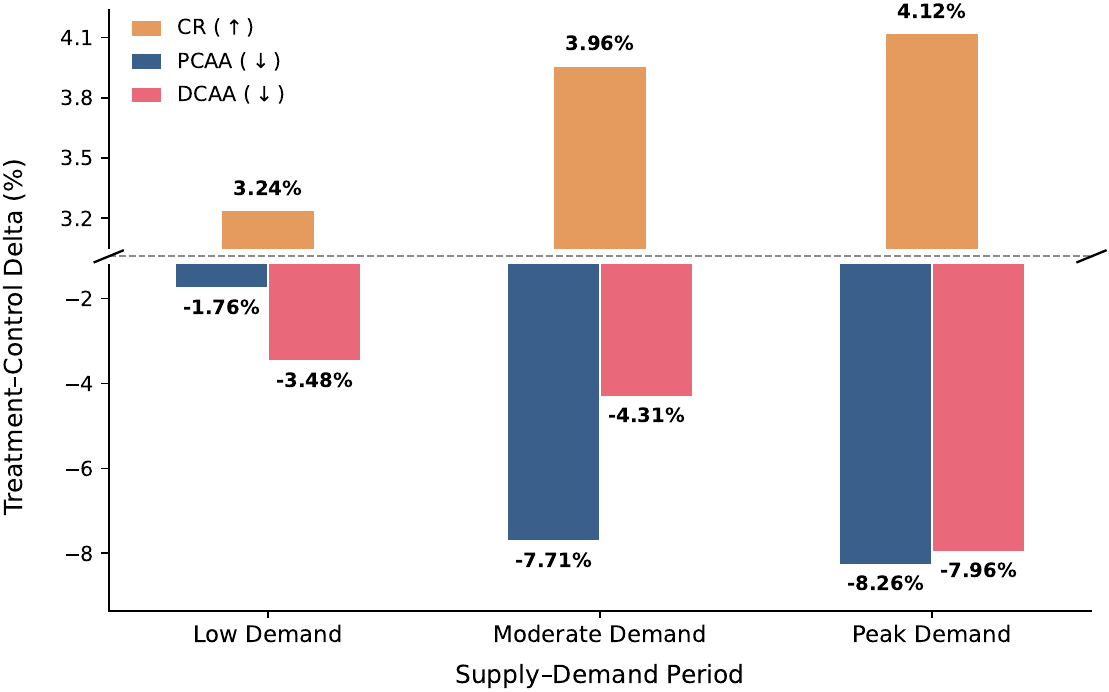}
  \caption{Online gains of GenMatch across supply--demand periods (T$-$C vs.\ PDP$_{\mathrm{KM}}$).}
  \Description{A grouped bar chart comparing low-, moderate-, and peak-demand periods. Completion-ratio gains increase from 3.24 percent to 3.96 percent and 4.12 percent. Passenger cancellation-after-answer ratio decreases by 1.76 percent, 7.71 percent, and 8.26 percent. Driver cancellation-after-answer ratio decreases by 3.48 percent, 4.31 percent, and 7.96 percent.}
  \label{fig:period_performance}
\end{figure}

\noindent\textbf{Performance across supply--demand periods.}
Figure~\ref{fig:period_performance} shows that GenMatch remains effective under different supply--demand conditions and delivers larger gains in busier periods. From low-demand to peak-demand periods, the CR improvement rises from 3.24\% to 4.12\%. The reductions in PCAA and DCAA also expand from 1.76\% and 3.48\% to 8.26\% and 7.96\%, respectively. This trend is particularly notable because the generation targets contain only OD pairs from rides that were actually completed online, as described in Section~\ref{sec:methods}. Unbroadcast candidates have no observed service-process labels and are excluded from auxiliary supervision, but remain part of the complete dispatch batch and the generative candidate set. GenMatch can therefore transfer the completion patterns learned from completed pairs to previously unbroadcast candidates with similar structural and business characteristics. This ability becomes more valuable in peak periods, where denser competition leaves more latent completion opportunities unexplored, and converts them into additional completed rides and business gains.

\begin{figure}[!htbp]
  \centering
  \includegraphics[width=\linewidth]{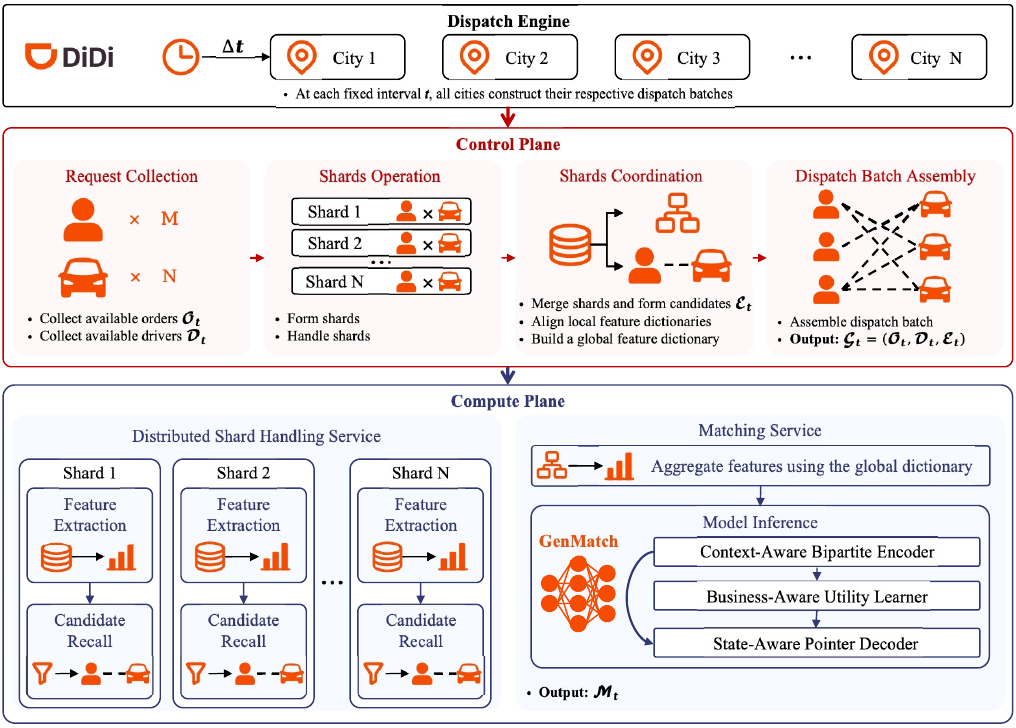}
  \caption{Production architecture of the Batch-Level Generative Dispatch Engine.}
  \Description{The production architecture contains a control plane that collects requests, coordinates distributed shards, and assembles dispatch batches, and a compute plane that performs distributed feature extraction, candidate retrieval, and GenMatch inference.}
\label{fig:dispatch_engine}
\end{figure}

\section{Deployment}
\label{sec:deployment}

The existing Pair-Level Dispatch Engine processes OD pairs independently and therefore cannot provide the complete dispatch batch required by GenMatch. We develop a Batch-Level Generative Dispatch Engine for production serving. As shown in Figure~\ref{fig:dispatch_engine}, its control plane coordinates distributed shards, restores a consistent candidate order, and assembles the complete batch. Its compute plane retains distributed feature extraction and candidate retrieval, then performs global GenMatch inference over the assembled batch. Generated pairs are treated as pre-assignments and enter the existing arbitration and locking process, preventing conflicts with other product lines. Malformed outputs, timeouts, or serving failures automatically fall back to the pair-level engine, allowing GenMatch to be deployed without weakening production reliability. Appendix~\ref{sec:appendix_deployment} provides the engineering details.

\section{Conclusion}

This paper presents GenMatch, an end-to-end generative framework deployed in a real-world production environment for MICOD. By directly generating the final assignment, GenMatch addresses the cross-stage objective inconsistency of the conventional multi-stage paradigm. Its Context-Aware Bipartite Encoder, Business-Aware Utility Learner, and State-Aware Pointer Decoder address the three challenges of encoding dynamic batch-level structures, learning unified business utility from heterogeneous feedback, and generating assignments under an evolving matching state, respectively. Extensive offline experiments validate these targeted designs, while online A/B tests demonstrate the effectiveness and practicality of GenMatch in production.

\bibliographystyle{ACM-Reference-Format}
\bibliography{sample-base}

\begin{table*}[!t]
  \centering
  \caption{Scale statistics of the five experimental cities.}
  \label{tab:city_stats}
  \setlength{\tabcolsep}{8pt}
  \renewcommand{\arraystretch}{1.15}
  \footnotesize
  \begin{tabular}{@{}ccccc@{}}
    \hline
    \textbf{Split} & \textbf{City} & \textbf{Daily Completed Orders}
    & \textbf{Daily Online Drivers} & \textbf{Avg.\ Trip Distance (m)} \\
    \hline
    \multirow{2}{*}{\textbf{Offline}} & City~I & 3.04e+03 & 0.18e+03 & 3.47e+03 \\
    & City~II & 1.12e+04 & 0.55e+03 & 4.41e+03 \\
    \hline
    \textbf{Offline \& Online} & City~III & 5.91e+03 & 0.43e+03 & 3.99e+03 \\
    \hline
    \multirow{2}{*}{\textbf{Online}} & City~IV & 1.73e+03 & 0.31e+03 & 4.29e+03 \\
    & City~V & 1.38e+04 & 0.68e+03 & 4.71e+03 \\
    \hline
  \end{tabular}
\end{table*}

\begin{table*}[!t]
  \centering
  \caption{Complete ablation results relative to GenMatch (Full). Values are the mean\,$\pm$\,standard deviation of percentage changes over five runs.}
  \label{tab:full_ablation_study}
  \setlength{\tabcolsep}{0.5pt}
  \renewcommand{\arraystretch}{1.1}
  \footnotesize
  \resizebox{\textwidth}{!}{%
  \begin{tabular}{@{}ll|cccc|cccc|cccc@{}}
    \hline
    \multirow{2}{*}{\textbf{Module}}
    & \multirow{2}{*}{\textbf{Variant}}
    & \multicolumn{4}{c|}{\textbf{City~I}}
    & \multicolumn{4}{c|}{\textbf{City~II}}
    & \multicolumn{4}{c}{\textbf{City~III}} \\
    & & \textbf{AR\,(\%)}\,$\uparrow$ & \textbf{CR\,(\%)}\,$\uparrow$ & \textbf{APT\,(\%)}\,$\downarrow$ & \textbf{GMV\,(\%)}\,$\uparrow$
    & \textbf{AR\,(\%)}\,$\uparrow$ & \textbf{CR\,(\%)}\,$\uparrow$ & \textbf{APT\,(\%)}\,$\downarrow$ & \textbf{GMV\,(\%)}\,$\uparrow$
    & \textbf{AR\,(\%)}\,$\uparrow$ & \textbf{CR\,(\%)}\,$\uparrow$ & \textbf{APT\,(\%)}\,$\downarrow$ & \textbf{GMV\,(\%)}\,$\uparrow$ \\
    \hline
    \textbf{GenMatch} & Full
    & \textbf{0.00\,$\pm$\,0.00} & \textbf{0.00\,$\pm$\,0.00} & \textbf{0.00\,$\pm$\,0.00} & \textbf{0.00\,$\pm$\,0.00}
    & \textbf{0.00\,$\pm$\,0.00} & \textbf{0.00\,$\pm$\,0.00} & \textbf{0.00\,$\pm$\,0.00} & \textbf{0.00\,$\pm$\,0.00}
    & \textbf{0.00\,$\pm$\,0.00} & \textbf{0.00\,$\pm$\,0.00} & \textbf{0.00\,$\pm$\,0.00} & \textbf{0.00\,$\pm$\,0.00} \\
    \hline
    \multirow{3}{*}{Encoder} & A1
    & $-$2.32\,$\pm$\,0.04 & $-$2.61\,$\pm$\,0.05 & $+$1.74\,$\pm$\,0.02 & $-$0.57\,$\pm$\,0.04
    & $-$2.14\,$\pm$\,0.07 & $-$2.55\,$\pm$\,0.04 & $+$1.23\,$\pm$\,0.04 & $-$2.07\,$\pm$\,0.10
    & $-$2.55\,$\pm$\,0.12 & $-$3.07\,$\pm$\,0.07 & $+$1.16\,$\pm$\,0.02 & $-$2.28\,$\pm$\,0.15 \\
    & A2
    & $-$2.25\,$\pm$\,0.02 & $-$2.30\,$\pm$\,0.02 & $+$1.61\,$\pm$\,0.01 & $-$0.52\,$\pm$\,0.02
    & $-$1.95\,$\pm$\,0.03 & $-$2.36\,$\pm$\,0.02 & $+$1.06\,$\pm$\,0.02 & $-$1.96\,$\pm$\,0.05
    & $-$2.43\,$\pm$\,0.06 & $-$2.69\,$\pm$\,0.03 & $+$1.14\,$\pm$\,0.01 & $-$2.15\,$\pm$\,0.06 \\
    & A3
    & $-$0.14\,$\pm$\,0.02 & $-$0.12\,$\pm$\,0.02 & $+$0.13\,$\pm$\,0.01 & $-$0.04\,$\pm$\,0.01
    & $-$0.66\,$\pm$\,0.05 & $-$1.04\,$\pm$\,0.04 & $+$0.49\,$\pm$\,0.03 & $-$0.72\,$\pm$\,0.10
    & $-$0.39\,$\pm$\,0.06 & $-$0.25\,$\pm$\,0.03 & $+$0.18\,$\pm$\,0.02 & $-$0.40\,$\pm$\,0.07 \\
    \hline
    \multirow{2}{*}{Learner} & A4
    & $-$2.47\,$\pm$\,0.05 & $-$2.69\,$\pm$\,0.06 & $+$2.03\,$\pm$\,0.01 & $-$0.59\,$\pm$\,0.05
    & $-$2.20\,$\pm$\,0.09 & $-$2.97\,$\pm$\,0.06 & $+$1.28\,$\pm$\,0.05 & $-$2.30\,$\pm$\,0.15
    & $-$2.76\,$\pm$\,0.16 & $-$3.15\,$\pm$\,0.08 & $+$1.17\,$\pm$\,0.03  & $-$2.32\,$\pm$\,0.18 \\
    & A5
    & $-$1.83\,$\pm$\,0.03 & $-$1.99\,$\pm$\,0.03 & $+$2.70\,$\pm$\,0.01 & $-$0.45\,$\pm$\,0.03
    & $-$1.61\,$\pm$\,0.05 & $-$1.90\,$\pm$\,0.03 & $+$1.44\,$\pm$\,0.03 & $-$1.70\,$\pm$\,0.08
    & $-$2.02\,$\pm$\,0.08 & $-$2.34\,$\pm$\,0.04 & $+$1.53\,$\pm$\,0.02 & $-$1.88\,$\pm$\,0.09 \\
    \hline
    \multirow{7}{*}{Decoder} & A6
    & $-$1.12\,$\pm$\,0.02 & $-$0.49\,$\pm$\,0.02 & $+$0.76\,$\pm$\,0.01 & $-$0.15\,$\pm$\,0.01
    & $-$0.98\,$\pm$\,0.04 & $-$0.90\,$\pm$\,0.02 & $+$0.53\,$\pm$\,0.02 & $-$0.52\,$\pm$\,0.03
    & $-$1.47\,$\pm$\,0.07 & $-$1.69\,$\pm$\,0.04 & $+$0.63\,$\pm$\,0.02 & $-$0.65\,$\pm$\,0.04 \\
    & A7
    & $-$0.22\,$\pm$\,0.01 & $-$0.18\,$\pm$\,0.01 & $+$0.20\,$\pm$\,0.00 & $-$0.06\,$\pm$\,0.01
    & $-$0.11\,$\pm$\,0.02 & $-$0.40\,$\pm$\,0.01 & $+$0.16\,$\pm$\,0.01 & $-$0.24\,$\pm$\,0.03
    & $-$0.37\,$\pm$\,0.05 & $-$0.28\,$\pm$\,0.02 & $+$0.17\,$\pm$\,0.01 & $-$0.43\,$\pm$\,0.04 \\
    & A8
    & $-$0.86\,$\pm$\,0.01 & $-$1.55\,$\pm$\,0.01 & $+$1.39\,$\pm$\,0.00 & $-$0.40\,$\pm$\,0.01
    & $-$0.67\,$\pm$\,0.02 & $-$1.31\,$\pm$\,0.01 & $+$0.90\,$\pm$\,0.01 & $-$1.37\,$\pm$\,0.04
    & $-$1.11\,$\pm$\,0.04 & $-$1.07\,$\pm$\,0.02 & $+$0.86\,$\pm$\,0.01 & $-$1.45\,$\pm$\,0.05 \\
    & A9
    & $-$0.92\,$\pm$\,0.03 & $-$0.63\,$\pm$\,0.04 & $+$0.58\,$\pm$\,0.02 & $-$0.27\,$\pm$\,0.04
    & $-$0.52\,$\pm$\,0.07 & $-$1.05\,$\pm$\,0.05 & $+$0.46\,$\pm$\,0.04 & $-$0.69\,$\pm$\,0.14
    & $-$1.22\,$\pm$\,0.14 & $-$0.96\,$\pm$\,0.09 & $+$0.48\,$\pm$\,0.04 & $-$0.78\,$\pm$\,0.19 \\
    & A10
    & $-$1.57\,$\pm$\,0.06 & $-$1.31\,$\pm$\,0.07 & $+$1.09\,$\pm$\,0.02 & $-$0.23\,$\pm$\,0.07
    & $-$1.35\,$\pm$\,0.11 & $-$1.88\,$\pm$\,0.07 & $+$0.71\,$\pm$\,0.06 & $-$0.82\,$\pm$\,0.20
    & $-$1.84\,$\pm$\,0.20 & $-$2.05\,$\pm$\,0.10 & $+$0.73\,$\pm$\,0.04 & $-$0.91\,$\pm$\,0.24 \\
    & A11
    & $-$0.46\,$\pm$\,0.07 & $-$0.85\,$\pm$\,0.08 & $+$0.50\,$\pm$\,0.03 & $-$0.31\,$\pm$\,0.09
    & $-$0.29\,$\pm$\,0.13 & $-$1.38\,$\pm$\,0.08 & $+$0.32\,$\pm$\,0.07 & $-$1.12\,$\pm$\,0.25
    & $-$0.79\,$\pm$\,0.24 & $-$0.84\,$\pm$\,0.12 & $+$0.42\,$\pm$\,0.05 & $-$1.04\,$\pm$\,0.30 \\
    & A12
    & $-$1.93\,$\pm$\,0.15 & $-$1.79\,$\pm$\,0.10 & $+$0.59\,$\pm$\,0.03 & $-$0.23\,$\pm$\,0.01
    & $-$1.08\,$\pm$\,0.11 & $-$1.17\,$\pm$\,0.14 & $+$0.25\,$\pm$\,0.03 & $-$1.09\,$\pm$\,0.07
    & $-$2.46\,$\pm$\,0.15 & $-$3.07\,$\pm$\,0.20 & $+$0.16\,$\pm$\,0.03 & $-$0.91\,$\pm$\,0.05 \\
    \hline
  \end{tabular}}
\end{table*}

\begin{table*}[!t]
  \centering
  \caption{Auxiliary prediction AUC relative to PDP, reported as the mean\,$\pm$\,standard deviation of percentage changes over five runs.}
  \label{tab:mtl_comparison_algorithm}
  \setlength{\tabcolsep}{4pt}
  \renewcommand{\arraystretch}{1.1}
  \footnotesize
  \resizebox{\textwidth}{!}{%
  \begin{tabular}{l|ccc|ccc|ccc}
    \hline
    \multirow{2}{*}{\textbf{Variant}}
    & \multicolumn{3}{c|}{\textbf{City~I}}
    & \multicolumn{3}{c|}{\textbf{City~II}}
    & \multicolumn{3}{c}{\textbf{City~III}} \\
    & $\mathbf{AUC}_{\mathrm{DA}}$\,(\%)\,$\uparrow$
    & $\mathbf{AUC}_{\mathrm{PCAA}}$\,(\%)\,$\uparrow$
    & $\mathbf{AUC}_{\mathrm{DCAA}}$\,(\%)\,$\uparrow$
    & $\mathbf{AUC}_{\mathrm{DA}}$\,(\%)\,$\uparrow$
    & $\mathbf{AUC}_{\mathrm{PCAA}}$\,(\%)\,$\uparrow$
    & $\mathbf{AUC}_{\mathrm{DCAA}}$\,(\%)\,$\uparrow$
    & $\mathbf{AUC}_{\mathrm{DA}}$\,(\%)\,$\uparrow$
    & $\mathbf{AUC}_{\mathrm{PCAA}}$\,(\%)\,$\uparrow$
    & $\mathbf{AUC}_{\mathrm{DCAA}}$\,(\%)\,$\uparrow$ \\
    \hline
    PDP
    & 0.00\,$\pm$\,0.00 & 0.00\,$\pm$\,0.00 & 0.00\,$\pm$\,0.00
    & 0.00\,$\pm$\,0.00 & 0.00\,$\pm$\,0.00 & 0.00\,$\pm$\,0.00
    & 0.00\,$\pm$\,0.00 & 0.00\,$\pm$\,0.00 & 0.00\,$\pm$\,0.00 \\
    $\mathbf{GenMatch}_{\mathrm{MTL}}$
    & \textbf{$+$1.62\,$\pm$\,0.11}
    & \textbf{$+$0.84\,$\pm$\,0.04}
    & \textbf{$+$0.74\,$\pm$\,0.07}
    & \textbf{$+$1.79\,$\pm$\,0.14}
    & \textbf{$+$0.76\,$\pm$\,0.05}
    & \textbf{$+$0.45\,$\pm$\,0.03}
    & \textbf{$+$1.74\,$\pm$\,0.15} & \textbf{$+$0.82\,$\pm$\,0.05} & \textbf{$+$0.55\,$\pm$\,0.03} \\
    \hline
  \end{tabular}}
\end{table*}

\appendix

\setcounter{topnumber}{4}
\setcounter{dbltopnumber}{4}
\setcounter{totalnumber}{6}
\renewcommand{\topfraction}{0.95}
\renewcommand{\dbltopfraction}{0.95}
\renewcommand{\textfraction}{0.05}
\renewcommand{\floatpagefraction}{0.85}
\renewcommand{\dblfloatpagefraction}{0.85}

\section{Additional Experimental Details}

\subsection{Experimental Details}
\label{sec:appendix_experimental_details}

\subsubsection{Datasets and City Statistics}
We evaluate GenMatch in five cities from DiDi's international ride-hailing markets. City~I and City~II are used for offline evaluation, City~IV and City~V are used for online evaluation, and City~III is included in both settings to connect the offline and online observations. As shown in Table~\ref{tab:city_stats}, the cities cover substantially different operating scales: daily completed orders range from approximately 1.7K to 13.8K, daily online drivers range from 0.18K to 0.68K, and average trip distance ranges from 3.47~km to 4.71~km. This diversity allows us to evaluate the method under different market sizes and supply--demand conditions rather than on a single operating environment.

For offline experiments, we construct city-specific simulation environments from historical dispatch logs. Fourteen days of data are used for training and seven days for evaluation. At every replayed dispatch step, the simulator provides the contemporaneous orders, available drivers, and feasible candidate connections. The upstream retrieval and feasibility-filtering system is fixed for all methods, so every method receives candidate sets generated under the same pickup-distance, service-range, and business constraints. Consequently, the comparison focuses on how each method represents, values, and selects from a common candidate set rather than on differences in candidate retrieval.

Each dispatch batch is represented as a sparse bipartite graph containing order nodes, driver nodes, and feasible OD-pair edges. An order is described by 113 features, a driver by 140 features, and a candidate edge by 32 features, giving 285 input fields in total. These fields cover spatiotemporal context, historical behavior, supply--demand statistics, and pair-specific business signals such as pickup cost and transaction value. The graph size changes across dispatch steps because the numbers of available orders, drivers, and feasible connections are determined by the current market state.

\subsubsection{Metrics}
We evaluate dispatch performance using Answer Ratio (AR), Completion Ratio (CR), Average Pickup Time (APT), and Gross Merchandise Volume (GMV). Together, these metrics describe whether an assignment is accepted, whether it is successfully fulfilled, how efficiently the driver reaches the passenger, and how much transaction value it creates.

\noindent\textbf{Answer Ratio (AR)} is the percentage of dispatched orders answered by drivers. A larger AR indicates that the selected OD pairs better match driver willingness and that fewer broadcasts are spent on assignments unlikely to receive a response. It therefore measures the immediate effectiveness of the dispatch decision at the first stage of the service process.

\noindent\textbf{Completion Ratio (CR)} is the percentage of assigned orders that ultimately complete their trips. Compared with AR, CR additionally reflects cancellations and other failures after an answer. A larger CR indicates that the generated assignments are more likely to survive the complete service process and become successfully fulfilled rides, making it a direct measure of dispatch reliability and service conversion.

\noindent\textbf{Average Pickup Time (APT)} measures the average elapsed time from assignment to driver pickup. A smaller APT means that drivers can reach passengers more quickly, reducing passenger waiting time and driver-side pickup cost. APT therefore captures the spatial and operational efficiency of the selected matching rather than only whether the trip is answered or completed.

\noindent\textbf{Gross Merchandise Volume (GMV)} is the total transaction value contributed by completed trips. A larger GMV indicates that the dispatch policy produces greater realized business value, jointly reflecting the number of completed rides and their transaction values.

No single metric fully characterizes dispatch quality. For example, aggressively prioritizing nearby candidates may reduce APT without maximizing completion or transaction value, while prioritizing high-value trips alone may increase pickup cost. We therefore assess the four metrics jointly: higher AR, CR, and GMV and lower APT indicate better overall performance.

\subsubsection{Baselines}
We compare GenMatch with production, end-to-end, value-based, and multi-agent dispatch methods. All methods operate on candidate sets produced by the same upstream retrieval and feasibility-filtering system.

\noindent\textbf{Production Dispatching Pipeline (PDP).}
PDP follows the conventional industrial paradigm: it first predicts business outcomes independently for each OD pair, combines these signals with a hand-crafted value function, and then constructs a batch-level assignment with a separate matching solver. PDP$_{\mathrm{KM}}$ is the deployed configuration and uses Kuhn--Munkres matching to optimize the assignment globally under the calculated pair weights. PDP$_{\mathrm{Greedy}}$ and PDP$_{\mathrm{GS}}$ retain the same upstream predictions and pair values but replace the final solver with greedy selection and Gale--Shapley matching, respectively. Comparing these variants isolates the effect of the matching solver within the conventional multi-stage pipeline.

\noindent\textbf{End-to-End MICOD Baseline.}
D2SN formulates MICOD as a two-layer Markov decision process and sequentially selects OD pairs or hold actions with an encoder--decoder reinforcement-learning policy. It is the most closely related end-to-end baseline because it also constructs an assignment sequentially. Unlike GenMatch, however, it does not explicitly model candidate interactions on the sparse bipartite graph or use direct multi-task supervision from stage-wise service outcomes.

\noindent\textbf{Value-Based Baselines.}
V1D3 and RLW improve dispatch decisions by learning or refining value estimates for candidate assignments. They represent approaches that enhance pair-level weights while retaining a value-driven decision pipeline. Their comparison with GenMatch tests whether improved pair values alone are sufficient, or whether jointly learning batch context and the final assignment provides additional benefit.

\noindent\textbf{Multi-Agent Baselines.}
CoRide and CoopRide model driver or regional cooperation with multi-agent reinforcement learning and are primarily designed to improve fleet-level, long-term supply--demand coordination. We include them to examine whether policies developed from a Macro-View perspective transfer effectively to the current-batch, OD-pair-level MICOD objective.

\noindent\textbf{GenMatch$_{\mathrm{Value}}$.}
This variant replaces only PDP's pair-level prediction stage with the DA, PCAA, and DCAA predictions learned by GenMatch. The original hand-crafted value calculation and Kuhn--Munkres matching stages remain unchanged. It isolates the benefit of the Business-Aware Utility Learner's service-process predictions, while the comparison between GenMatch$_{\mathrm{Value}}$ and full GenMatch measures the additional contribution of learned utility and state-aware assignment generation.

\subsubsection{Evaluation Protocols}
Offline evaluation replays seven days of historical dispatch traffic in each city-specific simulator after training on fourteen days of data. The replay preserves the observed order arrivals, driver states, and candidate connections, and all competing methods are evaluated under the same upstream feasibility rules. We report AR, CR, APT, and GMV relative to PDP$_{\mathrm{KM}}$. Each offline result is averaged over five independent runs, and the corresponding standard deviation reflects variation across those runs.

Online A/B tests are conducted in City~III--V for fourteen days. GenMatch and the PDP$_{\mathrm{KM}}$ control are evaluated with a one-hour time-slice interleaved design, and results are reported as treatment-minus-control changes. This design compares the two systems under recurring live traffic conditions while limiting long-term drift between the treatment and control periods. The online evaluation uses the same four primary dispatch metrics as the offline study and additionally examines broadcast volume, answer and completion counts, passenger cancellation and bad-experience ratios, driver answer and cancellation ratios, and driver income.

\subsubsection{Implementation Details}
All models are trained on four NVIDIA RTX L20 GPUs. We use DeepSpeed Zero Redundancy Optimizer (ZeRO) Stage~2 to partition optimizer states and gradients across devices and bfloat16 (BF16) mixed precision to reduce memory and computation costs. The default GenMatch configuration contains two sparse bipartite encoder layers and two pointer-decoder layers with hidden dimension 128. Matching attention, competition attention, and pointer scoring each use four heads, and the feed-forward dimension is 512. The model supports at most 500 orders, 500 drivers, and 10,000 candidate OD pairs in one dispatch batch.

Training uses Adam for 50 epochs, a cosine learning-rate schedule with three warm-up epochs, a learning-rate search range of $5\!\times\!10^{-5}$ to $5\!\times\!10^{-4}$, weight decay of $10^{-4}$, and gradient clipping at 1.0. The per-GPU batch size is 16 and the global batch size is 64. The three auxiliary service-process tasks receive equal task weights, and the overall multi-task-loss coefficient is set to $\lambda_{\mathrm{mtl}}=10$ according to the sensitivity analysis in Section~\ref{sec:appendix_loss_weight}. Offline experiments use five independent random seeds. Complete model and optimization settings are summarized in Table~\ref{tab:model_configuration}, and the training and inference procedures are provided in Section~\ref{sec:appendix_procedures}.

\subsection{Full Ablation Results}
\label{sec:appendix_full_ablation}
Table~\ref{tab:full_ablation_study} reports all twelve ablations on the three offline cities. We analyze each variant together with its corresponding result and implication, organized around the three modules of GenMatch.

\noindent\textbf{Context-Aware Bipartite Encoder (A1--A3).}
A1 removes both batch-level matching attention in Eq.~\eqref{eq:matching-encoding} and competition attention in Eq.~\eqref{eq:competition-encoding}, reducing the encoder to isolated OD-pair representations. This change degrades every metric, decreasing AR by 2.14\%--2.55\%, CR by 2.55\%--3.07\%, and GMV by 0.57\%--2.28\%, while increasing APT by 1.16\%--1.74\%. A2 restores matching attention over the complete candidate neighborhoods while still removing competition attention. Relative to A1, it improves AR by 0.07\%--0.19\%, CR by 0.19\%--0.38\%, and GMV by 0.05\%--0.13\%, while reducing the APT increase by 0.02\%--0.17\%; hence, neighborhood-level matching information is useful even without competition modeling. Full GenMatch further improves all metrics over A2, confirming that competition relations provide complementary batch context. A3 instead retains both attention mechanisms but removes their shared degree embedding. Its effect is small in City~I ($-$0.14\% AR and $-$0.12\% CR), becomes largest in City~II ($-$0.66\% AR, $-$1.04\% CR, and $-$0.72\% GMV), and is intermediate in City~III. This city-dependent degradation indicates that explicit neighborhood size is particularly useful in larger candidate graphs, where normalized attention alone cannot preserve the scale of local competition.

\noindent\textbf{Business-Aware Utility Learner (A4--A5).}
A4 removes the multi-task supervision of candidate memory while retaining the stage-wise predictions used to construct utility. It decreases AR by 2.20\%--2.76\%, CR by 2.69\%--3.15\%, and GMV by 0.59\%--2.32\%, producing the largest AR and CR degradations among the learner variants in every city. Thus, behavioral-outcome supervision improves not only the auxiliary predictions but also the shared candidate representations used for assignment. A5 retains this supervision but removes the utility logit from Eq.~\eqref{eq:pointer-distribution}, leaving the decoder to rely on its structural score. It degrades all four metrics and increases APT by 1.44\%--2.70\%, the largest APT increase among all ablations in every city. This result shows that the structural score alone cannot recover the service-efficiency trade-offs captured by explicit business-value guidance.

\noindent\textbf{State-Aware Pointer Decoder (A6--A12).}
A6, A7, and A8 remove the initial-, selected-, and residual-state representations, respectively, from the state-aware query in Eq.~\eqref{eq:state-query}. A7 causes the smallest degradation of the three, whereas A6 reduces AR by 0.98\%--1.47\% and A8 reduces CR by 1.07\%--1.55\% and GMV by 0.40\%--1.45\%. The contrast shows that the original batch and, especially, the remaining feasible opportunities provide more decision context than the already selected pairs, although all three summaries contribute. A9 removes progress encoding and consistently degrades every metric, including AR by 0.52\%--1.22\% and CR by 0.63\%--1.05\%. This confirms that the averages of the three state sets do not by themselves retain how far generation has progressed. A10 replaces the explicit state-aware query with causal self-attention; its AR and CR losses reach 1.84\% and 2.05\% in City~III, showing that implicit history propagation does not represent the evolving feasible set as effectively. A11 trains with a fixed target order and remains inferior to Full in every city, with CR decreasing by 0.84\%--1.38\% and GMV by 0.31\%--1.12\%; random target permutations therefore reduce dependence on an arbitrary generation order. Finally, A12 uses the first-step generation logits as fixed weights for Kuhn--Munkres matching. It produces particularly large AR and CR losses, reaching $-$2.46\% and $-$3.07\%, respectively, in City~III. Although its APT change is comparatively small, the loss in answered and completed orders demonstrates that frozen scores and one-shot optimization cannot replace the score updates required after each selected pair changes the feasible set.

Overall, every targeted removal degrades all four dispatch metrics across all three cities. The encoder results establish the complementary roles of neighborhood composition, competition context, and neighborhood size; the learner results validate behavioral supervision and explicit business guidance; and the decoder results show that assignment quality depends on dynamically representing and updating the matching state. These complete results support the conclusions drawn from the compact City~III table in the main text.

\subsection{Auxiliary Prediction Evaluation}
\label{sec:appendix_auxiliary_prediction}

We use the area under the receiver operating characteristic curve (AUC) to evaluate the auxiliary DA, PCAA, and DCAA predictions. A larger AUC indicates better discrimination between positive and negative outcomes. Table~\ref{tab:mtl_comparison_algorithm} reports relative AUC changes over the production prediction model.

GenMatch improves all three tasks in every city, demonstrating that the gains are consistent across different markets rather than being specific to one dataset. DA improves by 1.62\%--1.79\%, the largest gain among the three tasks. PCAA and DCAA improve by 0.76\%--0.84\% and 0.45\%--0.74\%, respectively. The relatively small standard deviations over five runs further indicate stable improvements.

These results verify that the Business-Aware Utility Learner captures the stage-wise service process more accurately than the production prediction model. They also explain the improvement of GenMatch$_{\mathrm{Value}}$: replacing only the prediction stage with these auxiliary predictions already benefits dispatch performance, even when the hand-crafted value calculation and Kuhn--Munkres matching remain unchanged. The additional improvement of full GenMatch reported in the main text therefore comes from jointly learning business utility and generating the final assignment, rather than from prediction accuracy alone.

\subsection{Effect of Model Capacity}

\begin{figure*}[!t]
  \centering
  \begin{subfigure}[t]{0.48\textwidth}
    \centering
    \includegraphics[width=\linewidth]{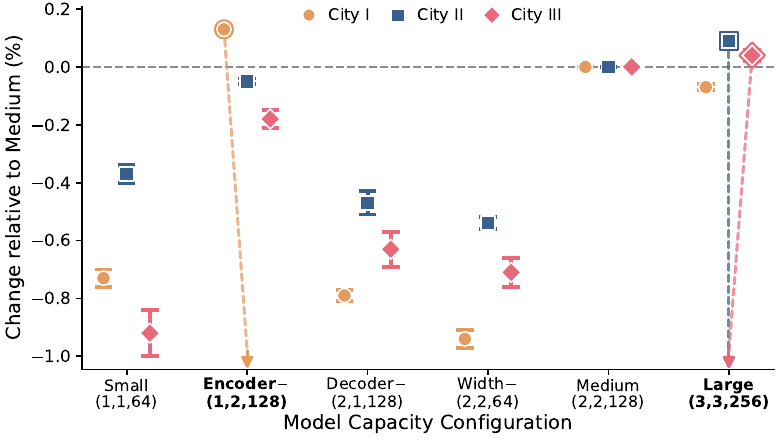}
    \caption{AR ($\uparrow$)}
    \label{fig:model_capacity_ar}
  \end{subfigure}
  \hfill
  \begin{subfigure}[t]{0.48\textwidth}
    \centering
    \includegraphics[width=\linewidth]{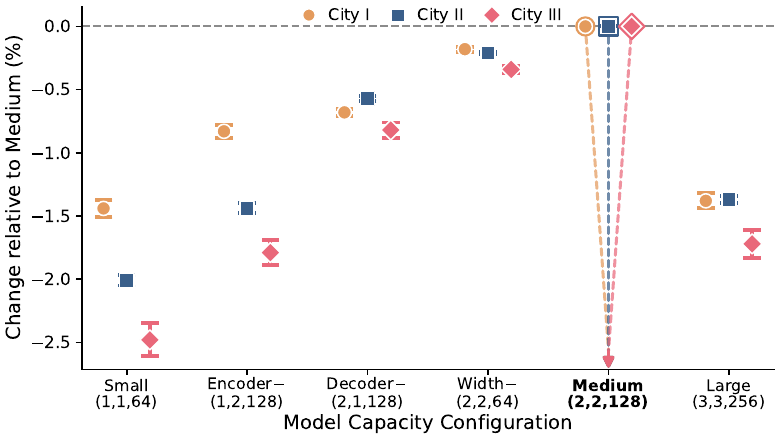}
    \caption{CR ($\uparrow$)}
    \label{fig:model_capacity_cr}
  \end{subfigure}

  \medskip
  \begin{subfigure}[t]{0.48\textwidth}
    \centering
    \includegraphics[width=\linewidth]{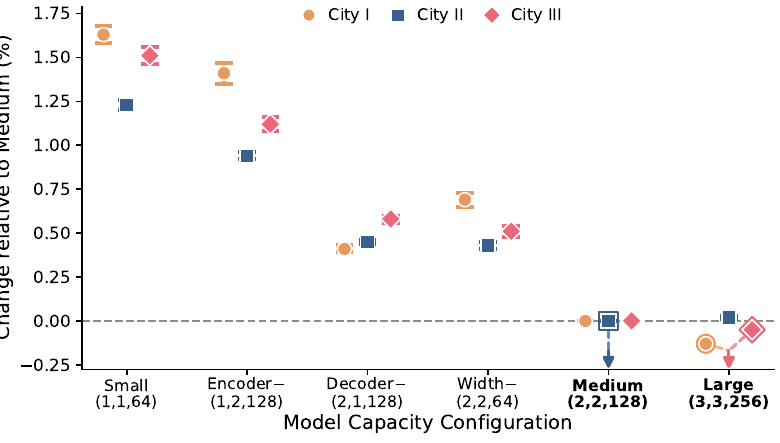}
    \caption{APT ($\downarrow$)}
    \label{fig:model_capacity_apt}
  \end{subfigure}
  \hfill
  \begin{subfigure}[t]{0.48\textwidth}
    \centering
    \includegraphics[width=\linewidth]{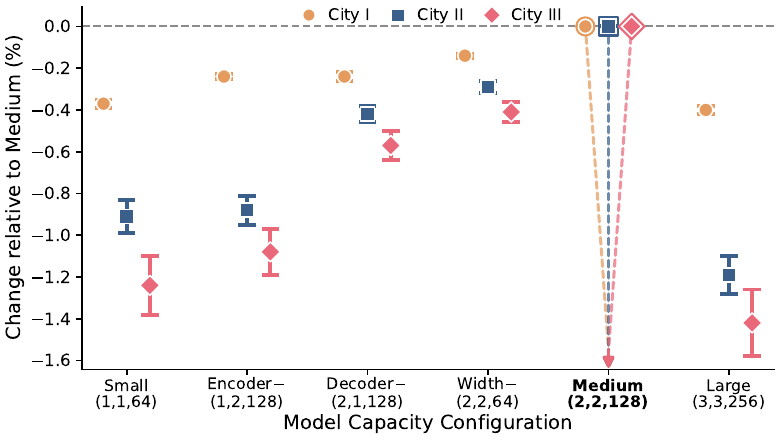}
    \caption{GMV ($\uparrow$)}
    \label{fig:model_capacity_gmv}
  \end{subfigure}
  \caption{Effect of model capacity relative to the Medium configuration. Error bars denote standard deviations over five runs.}
  \Description{Four panels show changes in answer ratio, completion ratio, average pickup time, and gross merchandise volume relative to the Medium configuration across three cities. Six configurations vary encoder depth, decoder depth, and hidden dimension.}
  \label{fig:model_capacity}
\end{figure*}

Figure~\ref{fig:model_capacity} studies model capacity by varying encoder depth, decoder depth, and hidden dimension $d$. Results are reported relative to the \textit{Medium} configuration, and the outlined markers identify the best configuration for each city. The \textit{Small} model degrades all four metrics across all three cities, showing that insufficient capacity limits both matching quality and business value. Reducing encoder depth, decoder depth, or width also causes broad performance drops. A shallower encoder markedly reduces CR and increases APT, confirming the importance of sufficient capacity for batch-level graph encoding. A shallower decoder mainly hurts AR and CR, while reduced width weakens all four metrics.

Increasing capacity beyond \textit{Medium} does not yield consistent gains. The \textit{Large} model slightly improves AR in City~II and City~III and reduces APT in City~I and City~III. However, it substantially reduces CR and GMV in every city. In contrast, \textit{Medium} achieves the best CR and GMV across all three cities and the best APT in City~II. It therefore provides the best overall balance between model capacity and dispatch performance and is used as the default configuration.

\captionsetup[subfigure]{
  font=small,
  labelfont=bf,
  justification=centering,
  skip=2pt
}

\subsection{Effect of the Multi-Task-Loss Weight}
\label{sec:appendix_loss_weight}

Table~\ref{tab:loss_weight} studies $\lambda_{\mathrm{mtl}}$, which balances the sequence-generation loss and the multi-task loss in Eq.~\eqref{eq:total-loss}. All values are reported relative to $\lambda_{\mathrm{mtl}}=10$. Weights below 10 provide insufficient behavioral supervision and consistently degrade dispatch performance. Larger weights can improve auxiliary AUC, especially at 1,000 and 10,000, but these prediction gains do not translate into better assignments: CR and GMV generally decrease because the multi-task objective begins to dominate sequence learning. We therefore set $\lambda_{\mathrm{mtl}}=10$, which provides the best overall balance across dispatch metrics and cities.

\begin{table*}[!t]
  \centering
  \caption{Sensitivity to $\lambda_{\mathrm{mtl}}$ relative to the selected value 10. Values are the mean\,$\pm$\,standard deviation of percentage changes over five runs. Larger AR, CR, GMV, and AUC and smaller APT are preferred. Bold and underlined values denote the best and second-best results in each column, respectively.}
  \label{tab:loss_weight}
  \setlength{\tabcolsep}{1.2pt}
  \renewcommand{\arraystretch}{1.08}
  \scriptsize
  \textbf{(a) Dispatch performance}\par\smallskip
  \resizebox{\textwidth}{!}{%
  \begin{tabular}{c|rrrr|rrrr|rrrr}
    \hline
    \multirow{2}{*}{$\lambda_{\mathrm{mtl}}$}
    & \multicolumn{4}{c|}{City~I} & \multicolumn{4}{c|}{City~II} & \multicolumn{4}{c}{City~III} \\
    & AR$\uparrow$ & CR$\uparrow$ & APT$\downarrow$ & GMV$\uparrow$
    & AR$\uparrow$ & CR$\uparrow$ & APT$\downarrow$ & GMV$\uparrow$
    & AR$\uparrow$ & CR$\uparrow$ & APT$\downarrow$ & GMV$\uparrow$ \\
    \hline
    0.01 & $-$0.29$\pm$0.05 & $-$0.44$\pm$0.06 & $+$0.67$\pm$0.01 & $-$0.19$\pm$0.05 & $-$0.27$\pm$0.05 & $-$0.29$\pm$0.06 & $+$0.43$\pm$0.01 & $-$0.57$\pm$0.05 & $-$0.78$\pm$0.05 & $-$0.88$\pm$0.06 & $+$0.22$\pm$0.01 & $-$0.60$\pm$0.05 \\
    0.1  & $-$0.17$\pm$0.02 & $-$0.39$\pm$0.02 & $+$0.37$\pm$0.01 & $-$0.08$\pm$0.02 & $-$0.19$\pm$0.02 & $-$0.22$\pm$0.02 & $+$0.24$\pm$0.01 & $-$0.24$\pm$0.02 & $-$0.49$\pm$0.02 & $-$0.46$\pm$0.02 & $+$0.16$\pm$0.01 & $-$0.35$\pm$0.02 \\
    1    & $-$0.11$\pm$0.03 & \underline{$-$0.07$\pm$0.03} & $+$0.20$\pm$0.01 & $-$0.02$\pm$0.03 & $-$0.08$\pm$0.03 & \underline{$-$0.14$\pm$0.03} & $+$0.12$\pm$0.01 & $-$0.18$\pm$0.03 & $-$0.20$\pm$0.03 & \underline{$-$0.17$\pm$0.03} & $+$0.05$\pm$0.01 & \underline{$-$0.11$\pm$0.03} \\
    10 & \underline{0.00$\pm$0.00} & \textbf{0.00$\pm$0.00} & 0.00$\pm$0.00 & \underline{0.00$\pm$0.00} & \textbf{0.00$\pm$0.00} & \textbf{0.00$\pm$0.00} & 0.00$\pm$0.00 & \textbf{0.00$\pm$0.00} & \underline{0.00$\pm$0.00} & \textbf{0.00$\pm$0.00} & 0.00$\pm$0.00 & \textbf{0.00$\pm$0.00} \\
    100  & \textbf{$+$0.21$\pm$0.05} & $-$0.11$\pm$0.06 & $+$0.13$\pm$0.01 & \textbf{$+$0.05$\pm$0.05} & \underline{$-$0.07$\pm$0.05} & $-$0.16$\pm$0.06 & $-$0.13$\pm$0.01 & \underline{$-$0.06$\pm$0.05} & \textbf{$+$0.08$\pm$0.05} & $-$0.33$\pm$0.06 & $-$0.08$\pm$0.01 & $-$0.28$\pm$0.05 \\
    1000 & $-$0.30$\pm$0.02 & $-$0.19$\pm$0.02 & \underline{$-$0.56$\pm$0.01} & $-$0.05$\pm$0.02 & $-$0.21$\pm$0.02 & $-$0.19$\pm$0.02 & \underline{$-$0.36$\pm$0.01} & $-$0.30$\pm$0.02 & $-$0.15$\pm$0.02 & $-$0.39$\pm$0.02 & \underline{$-$0.16$\pm$0.01} & $-$0.35$\pm$0.02 \\
    10000 & $-$0.67$\pm$0.03 & $-$0.25$\pm$0.03 & \textbf{$-$0.81$\pm$0.01} & $-$0.17$\pm$0.03 & $-$0.42$\pm$0.03 & $-$0.32$\pm$0.03 & \textbf{$-$0.47$\pm$0.01} & $-$0.67$\pm$0.03 & $-$0.49$\pm$0.03 & $-$0.44$\pm$0.03 & \textbf{$-$0.17$\pm$0.01} & $-$0.41$\pm$0.03 \\
    \hline
  \end{tabular}}

  \medskip
  \textbf{(b) Auxiliary prediction AUC}\par\smallskip
  \resizebox{0.82\textwidth}{!}{%
  \begin{tabular}{c|rrr|rrr|rrr}
    \hline
    \multirow{2}{*}{$\lambda_{\mathrm{mtl}}$}
    & \multicolumn{3}{c|}{City~I} & \multicolumn{3}{c|}{City~II} & \multicolumn{3}{c}{City~III} \\
    & DA & PCAA & DCAA & DA & PCAA & DCAA & DA & PCAA & DCAA \\
    \hline
    0.01 & $-$0.29$\pm$0.05 & $+$0.05$\pm$0.06 & $+$0.08$\pm$0.01 & $-$0.17$\pm$0.05 & $-$0.15$\pm$0.06 & $-$0.12$\pm$0.01 & $-$0.16$\pm$0.05 & $-$0.04$\pm$0.06 & $-$0.18$\pm$0.01 \\
    0.1  & $-$0.17$\pm$0.02 & $+$0.11$\pm$0.02 & $-$0.14$\pm$0.01 & $-$0.14$\pm$0.02 & $-$0.04$\pm$0.02 & $-$0.05$\pm$0.01 & $-$0.14$\pm$0.02 & $-$0.16$\pm$0.02 & $-$0.11$\pm$0.01 \\
    1    & $-$0.44$\pm$0.03 & $+$0.16$\pm$0.03 & $-$0.09$\pm$0.01 & $-$0.11$\pm$0.03 & $-$0.06$\pm$0.03 & $-$0.01$\pm$0.01 & $-$0.05$\pm$0.03 & $-$0.14$\pm$0.03 & $-$0.06$\pm$0.01 \\
    10 & 0.00$\pm$0.00 & 0.00$\pm$0.00 & 0.00$\pm$0.00 & 0.00$\pm$0.00 & 0.00$\pm$0.00 & 0.00$\pm$0.00 & 0.00$\pm$0.00 & 0.00$\pm$0.00 & 0.00$\pm$0.00 \\
    100  & $-$0.06$\pm$0.05 & $-$0.06$\pm$0.06 & $+$0.03$\pm$0.01 & \underline{$+$0.09$\pm$0.05} & $-$0.03$\pm$0.06 & $-$0.04$\pm$0.01 & $-$0.07$\pm$0.05 & $+$0.09$\pm$0.06 & $+$0.13$\pm$0.01 \\
    1000 & \underline{$+$0.13$\pm$0.02} & \underline{$+$0.18$\pm$0.02} & \underline{$+$0.15$\pm$0.01} & \textbf{$+$0.12$\pm$0.02} & \underline{$+$0.07$\pm$0.02} & \underline{$+$0.03$\pm$0.01} & \underline{$+$0.06$\pm$0.02} & \textbf{$+$0.18$\pm$0.02} & \textbf{$+$0.19$\pm$0.01} \\
    10000 & \textbf{$+$0.21$\pm$0.03} & \textbf{$+$0.24$\pm$0.03} & \textbf{$+$0.21$\pm$0.01} & $+$0.08$\pm$0.03 & \textbf{$+$0.14$\pm$0.03} & \textbf{$+$0.15$\pm$0.01} & \textbf{$+$0.12$\pm$0.03} & \underline{$+$0.15$\pm$0.03} & \underline{$+$0.18$\pm$0.01} \\
    \hline
  \end{tabular}}
\end{table*}

\section{Implementation and Deployment Details}

\subsection{Model Configuration}
\label{sec:appendix_model_config}

Table~\ref{tab:model_configuration} lists the model and training configurations of GenMatch.

\begin{table}[!ht]
  \centering
  \caption{Model and training configurations of GenMatch.}
  \label{tab:model_configuration}
  \setlength{\tabcolsep}{4pt}
  \renewcommand{\arraystretch}{1.1}
  \footnotesize
  \begin{tabular}{lcc}
    \hline
    \textbf{Configuration} & \textbf{Symbol} & \textbf{Value} \\
    \hline
    Encoder layers & $L_{\mathrm{enc}}$ & 2 \\
    Decoder layers & $L_{\mathrm{dec}}$ & 2 \\
    Hidden dimension & $d$ & 128 \\
    Matching-attention heads & -- & 4 \\
    Competition-attention heads & -- & 4 \\
    Pointer heads & $P_{\mathrm{ptr}}$ & 4 \\
    Feed-forward dimension & -- & 512 \\
    Dropout ratio & -- & 0.2 \\
    Multi-task shared-layer dimensions & -- & $[256,256]$ \\
    Multi-task tower dimensions & -- & $[256,128,64]$ \\
    DA loss weight & $\lambda_{\mathrm{DA}}$ & 1.0 \\
    PCAA loss weight & $\lambda_{\mathrm{PCAA}}$ & 1.0 \\
    DCAA loss weight & $\lambda_{\mathrm{DCAA}}$ & 1.0 \\
    Multi-task-loss weight & $\lambda_{\mathrm{mtl}}$ & 10.0 \\
    Maximum orders per batch & -- & 500 \\
    Maximum drivers per batch & -- & 500 \\
    Maximum candidate OD pairs & -- & 10000 \\
    Training epochs & -- & 50 \\
    Optimizer & -- & Adam \\
    Learning-rate range & -- & $5\!\times\!10^{-5}$--$5\!\times\!10^{-4}$ \\
    Learning-rate scheduler & -- & Cosine \\
    Warm-up epochs & -- & 3 \\
    Weight decay & -- & $10^{-4}$ \\
    Batch size per GPU & -- & 16 \\
    Global batch size & -- & 64 \\
    Gradient clipping & -- & 1.0 \\
    \hline
  \end{tabular}
\end{table}

\subsection{Training and Inference Procedures}
\label{sec:appendix_procedures}

Algorithm~\ref{alg:genmatch} summarizes training. Each dispatch batch is encoded once, after which the model learns stage-wise service outcomes and utility logits. A random permutation of the completed OD pairs provides the teacher-forced generation target, and the generation and auxiliary losses jointly update all model parameters.

\begin{algorithm}[!t]
\caption{GenMatch Training Procedure}
\label{alg:genmatch}
\begin{algorithmic}[1]
\Require Mini-batches of $\{\mathcal{G}_t,\mathcal{C}_t,
\{\mathbf{y}_{r},w_r\}_{r=1}^{P_t}\}$;
loss weights $\{\lambda_m\}_{m\in\mathcal{T}}$ and
$\lambda_{\mathrm{mtl}}$
\Ensure Trained model parameters $\Theta$
\State Initialize model parameters $\Theta$
\For{each training epoch}
    \For{each mini-batch $\mathcal{I}$}
        \State Initialize accumulated losses and valid-position counts
        \For{each dispatch batch $t\in\mathcal{I}$}
        \State Encode $\mathcal{G}_t$ as pair representations
        $\{\mathbf{z}_r\}_{r=1}^{P_t}$ using
        Eq.~\eqref{eq:pair-representation}
        \State Predict the DA, PCAA, and DCAA probabilities with the
        multi-task network
        \State Obtain $\{a_r\}_{r=1}^{P_t}$ using
        Eq.~\eqref{eq:business-logit}
        \State Uniformly sample an ordering of $\mathcal{C}_t$ to form
        $\mathcal{Y}_t^*=(e_{t,*}^{(1)},\ldots,e_{t,*}^{(K_t^*)})$,
        where $K_t^*=|\mathcal{C}_t|$
        \State Set $\mathbf{S}_{\mathrm{init}}\gets\sum_r\mathbf{z}_r$,
        $N_{\mathrm{init}}\gets P_t$, and
        $\mathbf{q}_{\mathrm{init}}\gets
        \mathbf{S}_{\mathrm{init}}/N_{\mathrm{init}}$
        \State Initialize the selected-state statistics with
        $(\mathbf{S}_{\mathrm{sel}}^{(1)},N_{\mathrm{sel}}^{(1)})
        \gets(\mathbf{0},0)$
        \State Initialize the residual-state statistics with
        $(\mathbf{S}_{\mathrm{res}}^{(1)},N_{\mathrm{res}}^{(1)})
        \gets(\mathbf{S}_{\mathrm{init}},P_t)$
        \State Set $m_r^{(1)}\gets1$ for all $r$
        \For{$k=1,\ldots,K_t^*$}
            \State Construct $\mathbf{q}^{(k)}$ using
            Eq.~\eqref{eq:state-query}
            \State Compute $\{\bar{s}_r^{(k)}\}$ and
            $\{p_r^{(k)}\}$ using Eqs.~\eqref{eq:pointer-score}
            and~\eqref{eq:pointer-distribution}
            \State Under teacher forcing, find $r_*^{(k)}$ such that
            $e_{r_*^{(k)}}=e_{t,*}^{(k)}$
            \State Accumulate $-\log p_{r_*^{(k)}}^{(k)}$
            \State Let $\mathcal{B}^{(k)}$ contain the target pair and
            its currently selectable conflicting pairs
            \State Update the selected- and residual-state statistics using
            Eq.~\eqref{eq:state-transition}, and mask $\mathcal{B}^{(k)}$
        \EndFor
        \EndFor
        \State Compute $\mathcal{L}_{\mathrm{gen}}$ and
        $\mathcal{L}_{\mathrm{mtl}}$ using
        Eqs.~\eqref{eq:generation-loss} and~\eqref{eq:mtl-loss}
        \State Compute $\mathcal{L}$ using Eq.~\eqref{eq:total-loss}
        \State Update $\Theta$ using $\nabla_{\Theta}\mathcal{L}$
    \EndFor
\EndFor
\State \Return $\Theta$
\end{algorithmic}
\end{algorithm}

Algorithm~\ref{alg:genmatch_inference} details inference. GenMatch first computes pair memory and utility logits for the complete dispatch batch. At each step, it constructs the query from the current matching state, greedily selects the highest-probability selectable candidate, masks every conflicting candidate, and incrementally updates the selected and residual states. This loop continues while $\sum_r m_r^{(k)}>0$, and the selected OD pairs form the final matching set.

\begin{algorithm}[!t]
\caption{GenMatch Inference Procedure}
\label{alg:genmatch_inference}
\begin{algorithmic}[1]
\Require Dispatch batch $\mathcal{G}_t$; trained model parameters $\Theta$
\Ensure Generated assignment $\mathcal{M}_t$
\State Encode $\mathcal{G}_t$ as pair representations
$\{\mathbf{z}_r\}_{r=1}^{P_t}$ using
Eq.~\eqref{eq:pair-representation}
\State Predict the DA, PCAA, and DCAA probabilities with the
multi-task network
\State Obtain $\{a_r\}_{r=1}^{P_t}$ using
Eq.~\eqref{eq:business-logit}
\State Initialize $\mathcal{M}_t\gets\varnothing$,
$m_r^{(1)}\gets1$ for all $r$, and $k\gets1$
\State Set $\mathbf{S}_{\mathrm{init}}\gets\sum_r\mathbf{z}_r$,
$N_{\mathrm{init}}\gets P_t$, and
$\mathbf{q}_{\mathrm{init}}\gets
\mathbf{S}_{\mathrm{init}}/N_{\mathrm{init}}$
\State Initialize the selected-state statistics as $(\mathbf{0},0)$
and the residual-state statistics as $(\mathbf{S}_{\mathrm{init}},P_t)$
\While{$\sum_{r=1}^{P_t}m_r^{(k)}>0$}
    \State Construct $\mathbf{q}^{(k)}$ using
    Eq.~\eqref{eq:state-query}
    \State Compute $\{\bar{s}_r^{(k)}\}$ and $\{p_r^{(k)}\}$ using
    Eqs.~\eqref{eq:pointer-score} and~\eqref{eq:pointer-distribution}
    \State Select $r^{(k)}$ using Eq.~\eqref{eq:greedy-selection}
    \State $\mathcal{M}_t\gets\mathcal{M}_t\cup\{e_{r^{(k)}}\}$
    \State Let $\mathcal{B}^{(k)}$ contain the selected pair and its
    currently selectable conflicting pairs
    \State Update the selected- and residual-state statistics using
    Eq.~\eqref{eq:state-transition}, and mask $\mathcal{B}^{(k)}$
    \State $k\gets k+1$
\EndWhile
\State \Return $\mathcal{M}_t$
\end{algorithmic}
\end{algorithm}

\subsection{Deployment Details}
\label{sec:appendix_deployment}
GenMatch has been deployed on a large-scale ride-hailing platform serving
multiple international markets. The existing production system, which we call
the Pair-Level Dispatch Engine, was designed for the conventional
multi-stage paradigm. It partitions feasible order-driver pairs into shards,
predicts pair-level business signals in parallel, calculates matching weights,
and constructs the final assignment. This design is efficient because each pair
can be processed independently before dispatch matching. It therefore avoids
assembling the complete dispatch batch during model inference and scales well
under strict latency constraints.

GenMatch changes the serving unit from an individual pair to an entire dispatch
batch. Its encoder requires the complete sparse bipartite graph, and its decoder
must maintain a consistent candidate order while generating assignments. We
therefore develop a Batch-Level Generative Dispatch Engine. This upgrade
introduces three challenges. First, centralizing feature preparation would cause
excessive compute and latency, while independent shards cannot provide the full
batch structure. We retain distributed candidate filtering and feature
preparation, then use a city-level orchestrator to restore candidate order and
assemble the complete batch for global inference. Second, GenMatch requires 285
features produced across different shards. Re-fetching them globally is costly,
and shard-local feature identifiers may be inconsistent. We pass sparse features
with their candidates and re-key them into one global feature dictionary during
aggregation. Third, autoregressive outputs depend on stable candidate indices
and must coexist with other product lines. We remove previously claimed
candidates before inference and treat generated pairs as pre-assignments, which
then enter the standard arbitration and locking process.

We deploy the new engine in three steps: feature transmission, shadow model
inference, and generative decision making. Any serving failure, malformed output,
or configuration error automatically falls back to the Pair-Level Dispatch
Engine, and GenMatch can be disabled without redeployment. The resulting design
retains the scalability of pair-level distributed processing while enabling safe
batch-level generative matching in production.

\end{document}